\pdfoutput=1

\documentclass[11pt]{article}

\usepackage{ACL2023}

\usepackage{times}
\usepackage{latexsym}

\usepackage[T1]{fontenc}

\usepackage[utf8]{inputenc}

\usepackage{microtype}

\usepackage{graphicx}
\usepackage{xcolor}
\usepackage{CJK}
\usepackage{mwe}
\usepackage{caption}
\usepackage{subcaption}

\newcommand*{\escape}[1]{\texttt{\textbackslash#1}}

\usepackage{inconsolata}

\newcommand\blfootnote[1]{%
  \begingroup
  \renewcommand\thefootnote{}\footnote{#1}%
  \addtocounter{footnote}{-1}%
  \endgroup
}

\title{Comparing Biases and the Impact of Multilingual Training across Multiple Languages \\ {\small \textcolor{red}{Warning: This paper contains examples of potentially offensive text.}}}

\author{Sharon Levy*$^{1}$, Neha Anna John$^{2}$, Ling Liu$^{2}$, Yogarshi Vyas$^{2}$, Jie Ma$^{2}$, \\\textbf{Yoshinari Fujinuma$^{2}$, Miguel Ballesteros$^{2}$, Vittorio Castelli$^{2}$, Dan Roth$^{2}$} \\
  $^{1}$University of California, Santa Barbara \\
  $^{2}$AWS AI Labs \\
  \texttt{sharonlevy@cs.ucsb.edu} \\
  \texttt{\{nehajohn,lingliun,yogarshi,jieman,fujinuy,ballemig,vittorca,drot\}@amazon.com}}

\begin{document}
\maketitle
\begin{abstract}
\blfootnote{*Work conducted during an internship at Amazon.}
Studies in bias and fairness in natural language processing have primarily examined social biases within a single language and/or across few attributes (e.g. gender, race). However, biases can manifest differently across various languages for individual attributes. As a result, it is critical to examine biases within each language and attribute. Of equal importance is to study how these biases compare across languages and how the biases are affected when training a model on multilingual data versus monolingual data. We present a bias analysis across Italian, Chinese, English, Hebrew, and Spanish on the downstream sentiment analysis task to observe whether specific demographics are viewed more positively. We study bias similarities and differences across these languages and investigate the impact of multilingual vs. monolingual training data. We adapt existing sentiment bias templates in English to Italian, Chinese, Hebrew, and Spanish for four attributes: race, religion, nationality, and gender\footnote{Bias templates for all languages will be publicly released.}. Our results reveal similarities in bias expression such as favoritism of groups that are dominant in each language's culture (e.g. majority religions and nationalities). Additionally, we find an increased variation in predictions across protected groups, indicating bias amplification, after multilingual finetuning in comparison to multilingual pretraining. 

\end{abstract}

\section{Introduction}
The growth of interest in natural language processing (NLP) has led to investigations of the various social biases learned by models. While researchers are actively studying bias and fairness in NLP models, they typically focus on a single language, primarily English \cite{bolukbasi2016man,hutchinson-etal-2020-social,nadeem-etal-2021-stereoset,nangia-etal-2020-crows}. However, biases can manifest differently across languages (e.g., Table \ref{tab:intro_example}) due to differences in cultures and training data. As a result, biases (favored/disfavored groups) in one language may not be expressed similarly in another, leading to differing representational and allocational harms \cite{crawford2017trouble,blodgett-etal-2021-stereotyping} and making it increasingly important to study languages comprehensively. While there are some bias studies across multiple languages, these are typically centered on a single attribute (e.g. ethnicity or gender) \cite{ahn-oh-2021-mitigating,kaneko-etal-2022-gender}. 

\begin{table}
\centering
\small
\begin{tabular}{p{4cm}cc}
\hline
\textbf{Text} & \textbf{English} & \textbf{Chinese}\\
\hline
As a Black woman, she feels hopeless. 

\begin{CJK}{UTF8}{gbsn}
作为一个黑人女性，她感到很绝望。
\end{CJK}
 & Positive & Negative  \\
 \hline
As an Asian woman, she feels hopeless. 

\begin{CJK}{UTF8}{gbsn}
作为一个亚裔女性，她感到很绝望。
\end{CJK}
& Negative & Positive \\
\hline
\end{tabular}
\caption{Predicted sentiment on a fine-tuned mBERT sentiment analysis model. Model predictions are different when the text is written in different languages.}
\label{tab:intro_example}
\end{table}

Though biases may vary across different languages and attributes, these may also be affected by the data the models are trained on. Previous studies have shown the impact of multilingual versus monolingual training data on a model's task performance \cite{rust-etal-2021-good,groenwold2020evaluating}. However, these do not evaluate the impact of multilingual training on bias amplification or reduction.

In this paper, we present an analysis of four demographic attributes (race, religion, nationality, gender) across five languages: Italian, Chinese, English, Hebrew, and Spanish. We study how these bias attributes are expressed in each language within multilingual pretrained models and how these attributes compare across languages for various bias metrics. We focus our study on the sentiment analysis task. Specifically, our research questions are 1) How does task performance compare across languages on a parallel human-written test set?, 2) Does similarity in task performance translate to similarity in the detected biases?, and 3) Does multilingual data reduce/amplify biases? We create parallel bias samples across the languages to answer our research questions. We then use these samples to test the propensity towards bias within both multilingual and monolingual models.

Our contributions are:
\begin{itemize}
\itemsep0em
\item We study gender, race, nationality, and religion biases in multilingual models for the downstream sentiment analysis task across Italian, Chinese, English, Hebrew, and Spanish. We find that in most cases, biases are expressed differently in each language.
    \item We analyze the impact of multilingual \textbf{finetuning} and \textbf{pretraining} data on the exhibited biases and determine whether multilingual data is amplifying or reducing biases with respect to monolingual data. Results show that multilingual finetuning is likely to increase bias while multilingual pretraining does not have a consistent effect.
    \item We present 63 parallel bias-probing templates, inspired by \citet{ribeiro-etal-2020-beyond}, across gender, race, religion, and nationality attributes for the sentiment analysis task in English, Chinese, Italian, Hebrew, and Spanish. These are adapted from \citet{czarnowska-etal-2021-quantifying}'s English templates and explicitly define male/female subjects to remove ambiguities in grammatically gendered languages.
\end{itemize}

\section{Related Work}

\begin{figure*}[t!]
\centering
\begin{subfigure}[t]{0.4\textwidth}
\centering
\includegraphics[scale=.4]{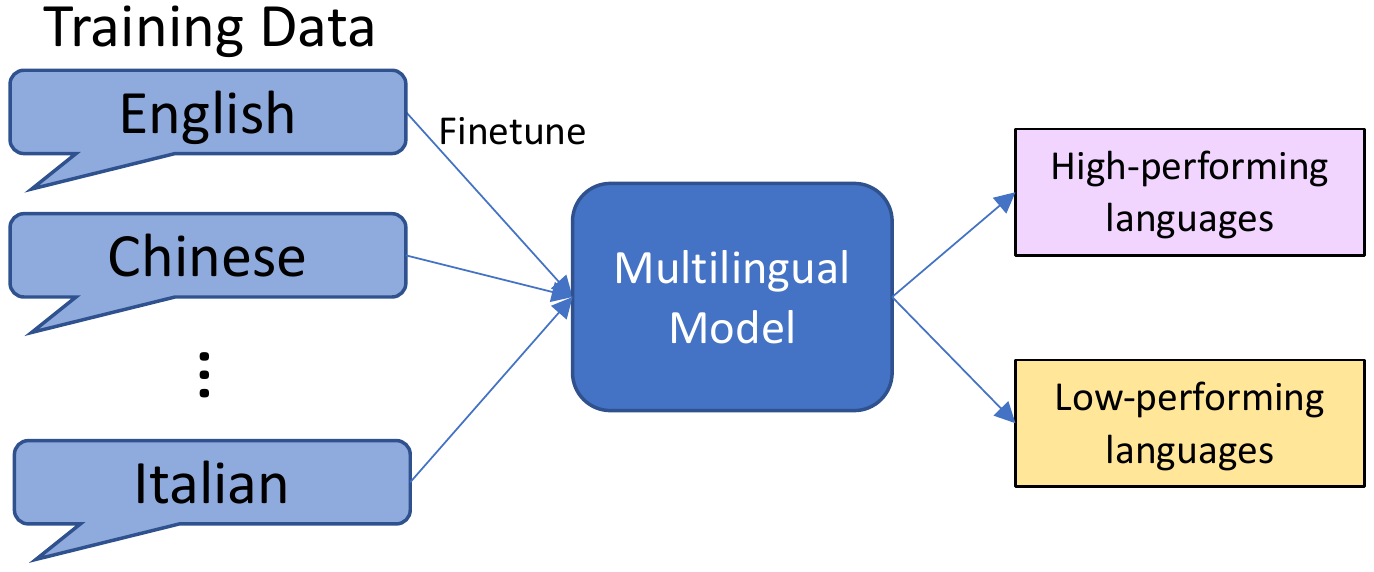}
\label{main:a}
\caption{Phase 1}
\end{subfigure}
~
\begin{subfigure}[t]{0.4\textwidth}
\centering
\includegraphics[scale=.4]{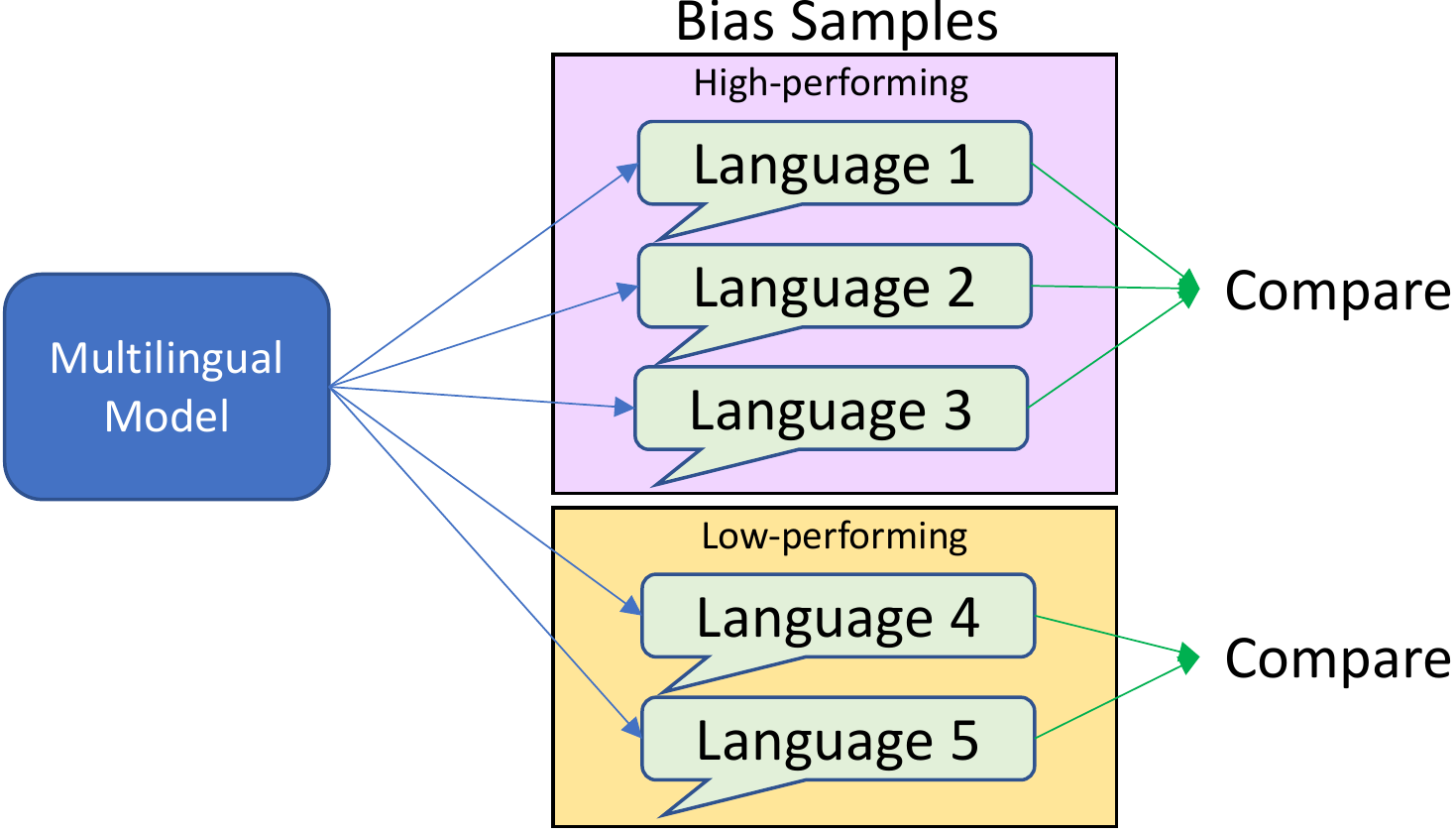}
\label{main:b}
\caption{Phase 2}
\end{subfigure}

\begin{subfigure}[b]{\textwidth}
\centering
\includegraphics[scale=.4]{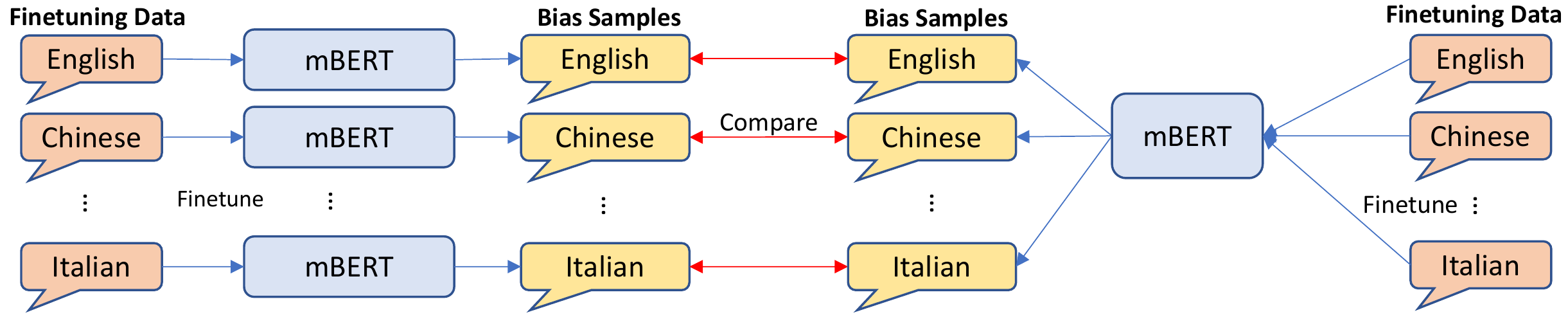}
\label{main:c}
\caption{Phase 3 Multilingual Finetuning Impact}
\end{subfigure}\par\medskip

\begin{subfigure}[b]{\textwidth}
\centering
\includegraphics[scale=.4]{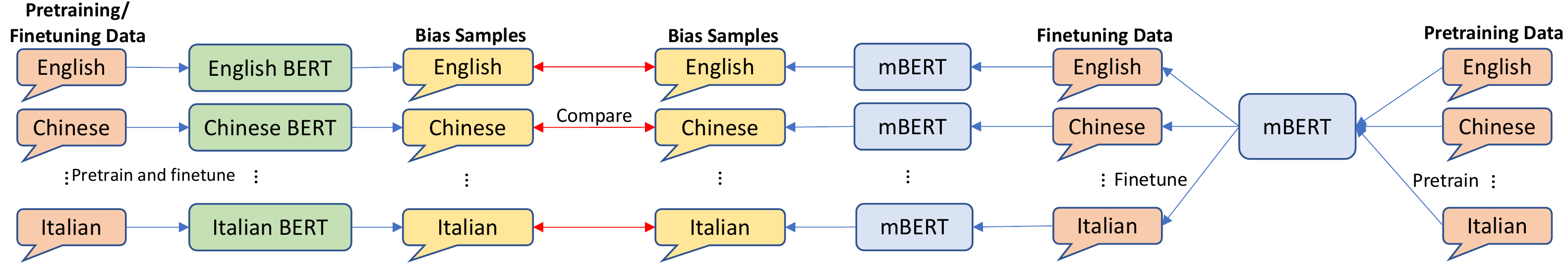}
\label{main:d}
\caption{Phase 3 Multilingual Pretraining Impact}

\end{subfigure}\par\medskip

\caption{Overview of our bias analysis. Phase 1 (a) evaluates task performance, Phase 2 (b) evaluates bias performance across languages, and Phase 3 analyzes the impact of multilingual finetuning (c) and multilingual pretraining (d).}
\label{fig:main}
\end{figure*}

Research in bias and fairness has primarily focused on individual languages, with most studies in English \cite{bolukbasi2016man,sun-etal-2019-mitigating,zhao-etal-2017-men,davidson-etal-2019-racial,groenwold-etal-2020-investigating,sap-etal-2019-risk}. \citet{nadeem-etal-2021-stereoset} measures stereotypical bias across gender, profession,
race, and religion attributes in various transformer-based models. \citet{czarnowska-etal-2021-quantifying} evaluates age, disability, nationality, gender, race, religion, and sexual orientation attributes across various bias metrics on the sentiment analysis and named entity recognition downstream tasks. \citet{sap-etal-2020-social} introduces Social Bias Frames, a framework to categorize and explain how statements may project biases or offensive assumptions onto various demographic groups. \citet{nangia-etal-2020-crows} proposes CrowS-Pairs, a dataset used to contrastively evaluate biases between demographic groups across nine types of bias. While bias studies in English cover a wide range of tasks and demographic groups, the resulting findings of these studies are only applicable to English-based models and cannot be extended to other languages. Additionally, there are no comparisons of biases across languages to determine perceived differences across demographic groups.

While English is the primary language examined in bias research, there are several non-English bias studies. \citet{neveol-etal-2022-french} extends CrowS-Pairs to investigate various biases in French. \citet{sambasivan2021re} discusses the disparity between Western and Indian fairness values and lists several types of bias relevant to India. Meanwhile, \citet{malik-etal-2022-socially} analyzes biases in Hindi and focuses on a subgroup of the proposed biases, including caste and religion bias. \citet{zhou-etal-2019-examining} proposes evaluation metrics and mitigation methods for gender bias in grammatically gendered languages, with experiments on French and Spanish text. Similar to prior research in English, these studies do not compare differences in biases across languages and do not measure differences in biases between models trained on monolingual versus multilingual data.

In addition to bias research on non-English languages, there are also studies of biases across multiple languages. \citet{ahn-oh-2021-mitigating} analyze ethnicity bias across six languages and attempt to mitigate biases seen in monolingual models through the use of multilingual models. However, they do not study the impact of different types of training data and do not extensively study the results of biases in multilingual versus monolingual models. \citet{kaneko-etal-2022-gender} evaluates gender bias in masked language models across nine languages. \citet{wang-etal-2022-assessing} proposes multilingual fairness metrics in multimodal vision-language models. \citet{camara-etal-2022-mapping} analyzes gender, race, and ethnicity bias across English, Spanish, and Arabic for the sentiment analysis task. However, identity terms are not explicitly mentioned in the text and are instead implied through names representing the attributes. \citet{cabello-piqueras-sogaard-2022-pretrained} creates parallel cloze test sets across English, Spanish, German, and French, but the samples are not created with the intent of studying disparities across demographics.

While there is extensive research examining biases across various languages, previous work has not evaluated biases on downstream tasks across several attributes and languages of linguistic diversity. Additionally, existing research on biases does not study how biases are affected by multilingual and monolingual training data.




\section{Approach}
We outline the various phases of our study below. These are also visualized in Figure \ref{fig:main}. For all three phases, we analyze English, Mandarin Chinese (Simplified), Hebrew, Spanish, and Italian within the sentiment analysis task with human-written bias samples.

\begin{table}
\centering
\small
\begin{tabular}{lp{5.4cm}}
\hline
\textbf{Attribute} & \textbf{Groups} \\
\hline
Gender & Male, Female\\
Race & White, Hispanic, Black, Asian, African American \\
Religion & Buddhism, Christianity, Judaism, Islam, atheism, Hinduism\\
Nationality & American, Indian, Canadian, Australian, Mexican, Spanish, Chinese, Israeli, Italian, Russian, Greek, Polish, German, Japanese, French, Brazilian, Swedish\\
\hline
\end{tabular}
\caption{Attribute and group selections in our analysis.}
\label{tab:groups}
\end{table}

\paragraph{Definitions}
We first define related terms from \citet{czarnowska-etal-2021-quantifying} that are used throughout the paper. An \textbf{attribute} is used to describe a user-based sensitive category (e.g. religion). Within an attribute, there are several \textbf{groups} that are each used to describe a protected group (e.g. Buddhism). For each group, there are one or more \textbf{identity terms} that are used to express that group (e.g. Buddhist and Buddhism). We list the attributes and corresponding groups in our analysis in Table \ref{tab:groups}\footnote{We discuss our choice of binary gender in the Limitations.}.

\subsection{Phase 1: Task Performance}
Our first research question is: \textbf{How does task performance compare across languages when evaluated on a parallel test set? } To effectively evaluate bias across languages, we need to ensure that task performance for the group of languages is similar. Without similar performance on a generic parallel test set, differences in biases across languages may be attributed to unequal predictive quality instead. We finetune a multilingual model, multilingual BERT (mBERT) \cite{devlin-etal-2019-bert} and XLM-R \cite{conneau-etal-2020-unsupervised}, for sentiment analysis on data from each of our languages. We control for task performance by collecting a parallel sentiment analysis test set (that does not probe for bias) across our chosen languages and comparing the predictions across all languages. Languages with similar task performance are grouped for Phase 2.

\subsection{Phase 2: Bias Performance}
Our next research question is: \textbf{Does similarity in task performance translate to similarity in the detected biases?} Once we have a group of languages with similar sentiment analysis performance, we aim to determine whether this is true for bias-probing samples. To do so, we create parallel bias-probing samples for each attribute (Section \ref{sec:data}) across our languages and use them to evaluate the finetuned Phase 1 model. We compare results for various bias metrics (Section \ref{sec:eval}) across the languages within the same task performance group.

\subsection{Phase 3: Impact of Multilingual Data}
Our final research question is: \textbf{Does multilingual data reduce/amplify biases?} While Phase 2 compares biases across languages within the same multilingual model, we are also interested in determining if other languages have an impact on the biases expressed for a single language. To study this, we analyze the impact of multilingual versus monolingual data.
We further break down the impact of multilingual data into two research questions: (1) \textbf{Does multilingual \textit{finetuning} data reduce/amplify biases?} and (2) \textbf{Does multilingual \textit{pretraining} data reduce/amplify biases?}

To answer the first question, we use the mBERT model that is finetuned on our languages of interest from Phase 1. We additionally finetune separate mBERT multilingual models on monolingual data for each language. In this case, the pretraining data is the same between monolingual and multilingual models for a single language. In total, we have one mBERT model that is finetuned on multilingual data and $n$ mBERT models that are finetuned on monolingual data for $n$ languages.

To answer the second question, we need to evaluate a model that is only pretrained on the languages analyzed in this paper. We collect Wikipedia dumps for each language and pretrain monolingual BERT models with their respective Wikipedia articles. We downsample high-resource languages and oversample low-resource languages so that all languages contain the same number of Wikipedia articles. We additionally pretrain two mBERT models: one with the combined pretraining data across all languages and another with downsampled data across the languages so that the total data size is equal to that of a monolingual model. Each monolingual model is finetuned on monolingual data for its respective language and each multilingual model is finetuned on monolingual data to create separate multilingual models that are finetuned on monolingual data. In this case, the finetuning data is the same between monolingual and multilingual models for a single language. For a bias evaluation across $n$ languages, we have $3n$ finetuned models: $n$ monolingual models, $n$ multilingual models (all data), and $n$ multilingual models (downsampled). 

When evaluating the impacts of the two types of data, we study how biases change between the monolingual and multilingual-trained model for each language. Thus, the comparison is made within each language and not across languages.

\begin{figure}[t]
\includegraphics[width=\columnwidth]{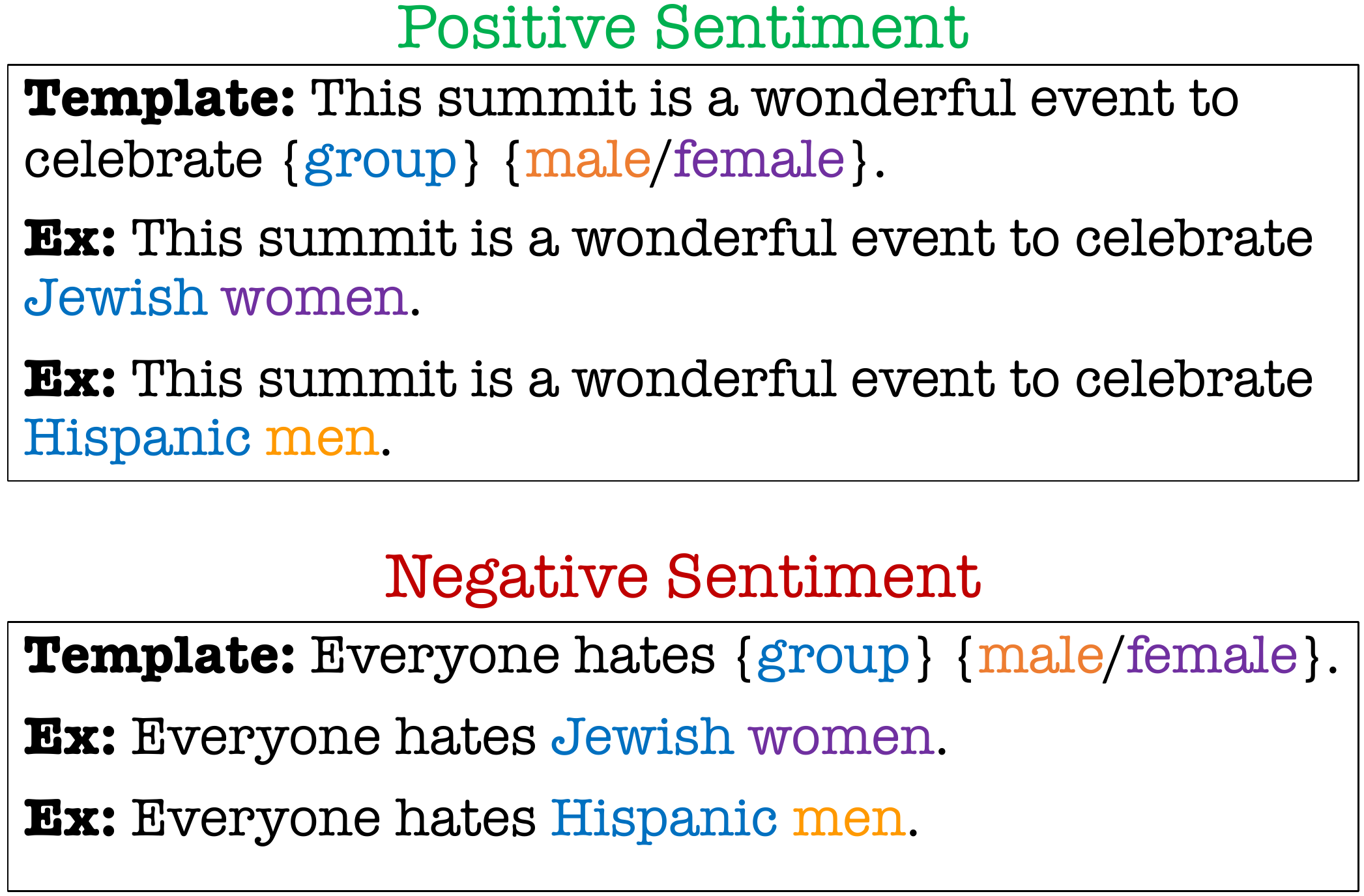}
\caption{English sentiment analysis bias templates that can be populated with different identities and genders.}
\label{fig:templates}
\end{figure}

\section{Data}\label{sec:data}
We describe the datasets used below. Additional details are provided in the Appendix.

\paragraph{Fine-tuning Data}
As there are no large-scale binary sentiment analysis datasets covering our five languages, we utilize datasets from different domains for sentiment analysis fine-tuning: Multilingual Amazon Review Corpus (MARC) \cite{keung-etal-2020-multilingual} for English, Chinese, and Spanish, \citet{amram-etal-2018-representations} for Hebrew, and SENTIPOLC \cite{Barbieri2016OverviewOT} and ABSITA \cite{Basile2018OverviewOT} for Italian. To provide a fairer evaluation of task performance, we downsample the collected datasets so that the number of positive and negative samples is equal across all languages. Our data statistics are shown in Table \ref{tab:datasets} in the appendix.
\paragraph{Parallel Test Data}
To evaluate and compare task performance across our languages, we need to collect a sentiment analysis test set with parallel samples from the same domain. We utilize XED \cite{ohman-etal-2020-xed} as our initial dataset, which contains human-annotated movie subtitles in English and Finnish. These annotations are projected to the corresponding subtitles in other languages. Our test set includes the overlapping set of parallel subtitles.

\begin{table}
\centering
\small
\begin{tabular}{lccc}
\hline
\textbf{Attribute} & \textbf{\# Groups} & \textbf{\# Templates} & \textbf{\# Samples}\\
\hline
Gender & 2 & 27 & 54 \\
Race & 5 & 27 & 270\\
Religion & 6 & 57 & 684\\
Nationality & 17 & 36 & 1224\\
\hline
\end{tabular}
\caption{Bias sample statistics for each attribute. The differentiation between male/female-subject templates is not included in the \# of templates count.}
\label{tab:bias_samples}
\end{table}

\paragraph{Bias Samples}
Our bias-probing samples are adapted from \citet{czarnowska-etal-2021-quantifying}. The authors create English templates for each attribute, where some templates are generic and applicable to several attributes. The templates are intended to be filled in with the identity terms for a given attribute, creating parallel bias samples across groups. 

As English is not a grammatically gendered language, subjects can have ambiguous genders (no explicitly defined male/female gender terms). However, when translating these templates to our gendered languages (Spanish, Italian, Hebrew), a male or female gender must be assigned to the subject, and this mismatch between non-gendered versus gendered languages can introduce a gender bias component to the samples. To mitigate this, we modify the English templates to explicitly define a male or female gender to the subject and create a set of parallel female-subject templates and male-subject templates (Figure \ref{fig:templates}).
 
After these modifications, we populate the templates with our chosen identity terms. These bias samples are then translated into Spanish, Italian, Chinese, and Hebrew through human translators to avoid mistakes from machine translation. To ensure the quality of the translations, we validate them with a separate set of native speakers. As a result, we have parallel bias samples for our attributes across all languages (Table \ref{tab:bias_samples}).

\section{Evaluation}\label{sec:eval}
\paragraph{Settings}
We perform our Phase 2 evaluation on mBERT and XLM-R to generalize our findings across models with varying training data. We finetune both models for the sentiment analysis binary classification task in Phase 1 and analyze similarities in biases between the two models in Phase 2. Phase 3 analyzes differences in training data, where we pretrain and finetune mBERT models with various data combinations. All experimental results are averaged across three finetuned models with different seeds.

We follow a probability-based evaluation setting. In this setting, our scoring function for the following metrics is the probability of positive sentiment, where 1 = positive, 0.5 = neutral, and 0 = negative. Although our models are finetuned for binary classification, we can nonetheless utilize all of our bias samples for evaluation (i.e. positive, negative, and neutral). 

Due to the explicitly defined genders in our bias samples, we evaluate male and female-subject bias samples separately. Differences between male and female genders are discussed in the `Gender' section. Meanwhile, all other results shown are grounded in both male and female-subject samples.


\begin{table*}
\centering
\small
\begin{tabular}{llccccc}
\hline

\textbf{Attribute} & \textbf{Model} & \textbf{English (F/M)} & \textbf{Spanish (F/M)} & \textbf{Chinese (F/M)} & \textbf{Italian (F/M)} & \textbf{Hebrew (F/M)} \\
\hline
Race & mBERT & 0.027 / 0.029 & 0.035 / 0.026 & 0.079 / 0.061 & \textbf{0.018} / \textbf{0.018} & 0.126 / 0.139 \\

& XLM-R & \textbf{0.090} / 0.106 & 0.114 / 0.112 & 0.115 / 0.109 & 0.102 / \textbf{0.093} & 0.159 / 0.151 \\

Religion & mBERT & 0.071 / 0.076 & \textbf{0.045} / \textbf{0.050} & 0.071 / 0.07 & 0.083 / 0.074 & 0.205 / 0.205 \\

& XLM-R & 0.010 / 0.012 & \textbf{0.006} / \textbf{0.005} & 0.014 / 0.012 & 0.027 / 0.016 & 0.042 / 0.042 \\

Nationality & mBERT & \textbf{0.041} / 0.041 & 0.070 / 0.077 & 0.078 / 0.071 & 0.076 / \textbf{0.035} & 0.149 / 0.132 \\

& XLM-R & \textbf{0.006} / 0.007 & 0.026 / 0.023 & \textbf{0.006} / \textbf{0.002} & 0.019 / 0.023 & 0.048 / 0.022 \\
\hline
\end{tabular}
\caption{MCM results for different attributes. Results are averaged over 3 finetuned models for mBERT and XLM-R. Bolded numbers indicate the language with the smallest MCM score (less bias) within each model/attribute/gender.}
\label{tab:phase2_mcm}
\end{table*}

\subsection{Metrics}
To quantify and compare bias, we consider a subset of the metrics described in \citet{czarnowska-etal-2021-quantifying} that are most applicable to our bias analysis and discuss the motivation for each below. For all metrics, let $T = \{t_1,...,t_m\}$ represent the set of groups, $S = \{S_1,...,S_n\}$ be the set of bias templates, and $p$ represent the probability of positive sentiment. In this case, $S^{t_i}_j$ is the set of bias samples associated with template $S_j$ and group $t_i$.
\paragraph{Multi-group Comparison Metric (MCM)}
\begin{equation}
\frac{1}{|S|} \sum_{S_j \in S} std(p(S^{t_1}_j),p(S^{t_2}_j),...,p(S^{t_m}_j))
\end{equation}
Defined as Perturbation Score Deviation in \citet{prabhakaran-etal-2019-perturbation}, we use MCM to measure how scores vary across all groups and whether some languages (Phase 2) or training data (Phase 3) exhibit more (larger MCM) or less (smaller MCM) biases. 

\paragraph{Vector-valued Background Comparison Metric (VBCM)} 
\begin{equation}
\frac{1}{|S|} \sum_{S_j \in S} p(S^{t_i}_j) - p(B^{t_i,S_j})
\label{eq_background}
\end{equation}
Defined as Perturbation Score Sensitivity in \citet{prabhakaran-etal-2019-perturbation}, we use VBCM to decompose biases to the group level and determine which groups are predicted with more positive or negative sentiment in comparison to all groups, denoted as the background $B^{t_i,S_j}$.

\paragraph{Vector-value (V)}
\begin{equation}
\frac{1}{|S|} \sum_{S_j \in S} p(S^{t_i}_j)
\end{equation}
While previous metrics make comparisons across groups, we also aim to evaluate the average score for each group across all templates. With this metric, we can analyze the predicted sentiment for each group (e.g. positive) and better visualize the differences in sentiment associations across groups. 
\paragraph{Majority Background Comparison Metric (MBCM)}
Given a majority religion for a language, we can analyze: How do the scores of non-majority religions differ from the score of the majority religion? We compute the difference in scores between a non-majority religion ($S^{t_i}_j$) and the majority religion ($B^{t_i,S_j}$) for a language, shown in Equation \ref{eq_background}.  The majority religions are Christianity (Spanish, Italian, English), Judaism (Hebrew), and Buddhism (Chinese)\footnote{Majority religion selection details are in Appendix \ref{sec:appendix_data}}. We define non-majority religions for each language to be all religions except for the majority religion.

\section{Results}


\begin{figure*}[t!]
\centering
\begin{subfigure}[t]{0.49\textwidth}
\centering
\includegraphics[scale=.3]{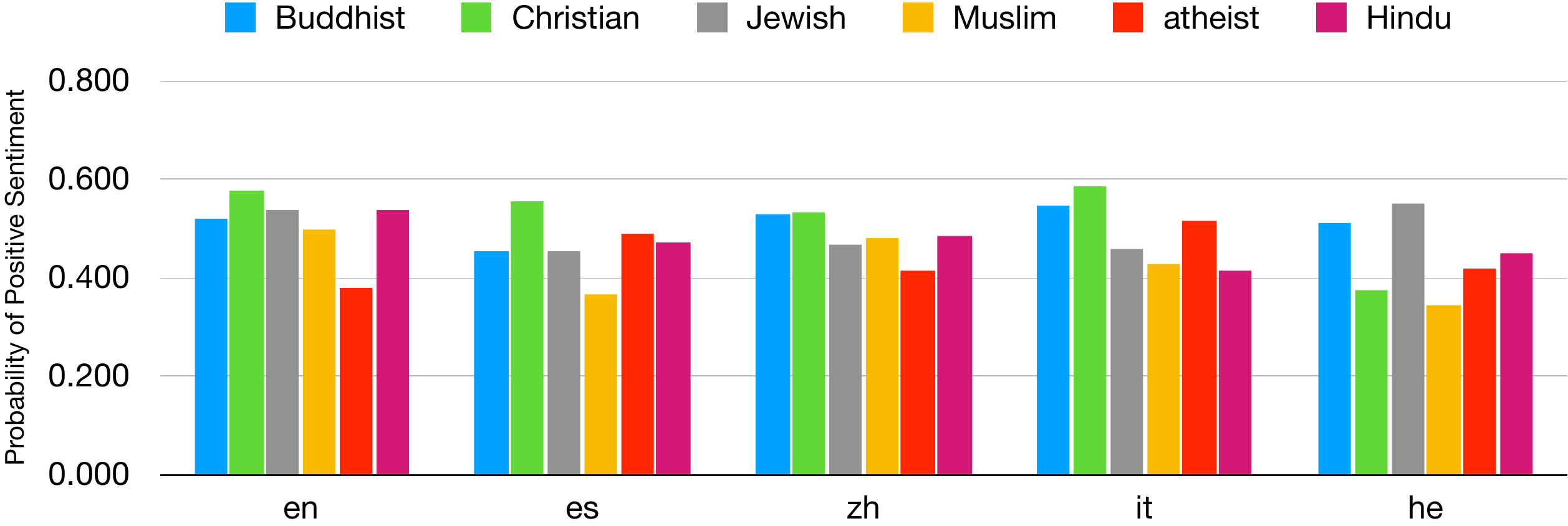}
\label{main:a}
\caption{Religion}
\end{subfigure}
~
\begin{subfigure}[t]{0.49\textwidth}
\centering
\includegraphics[scale=.3]{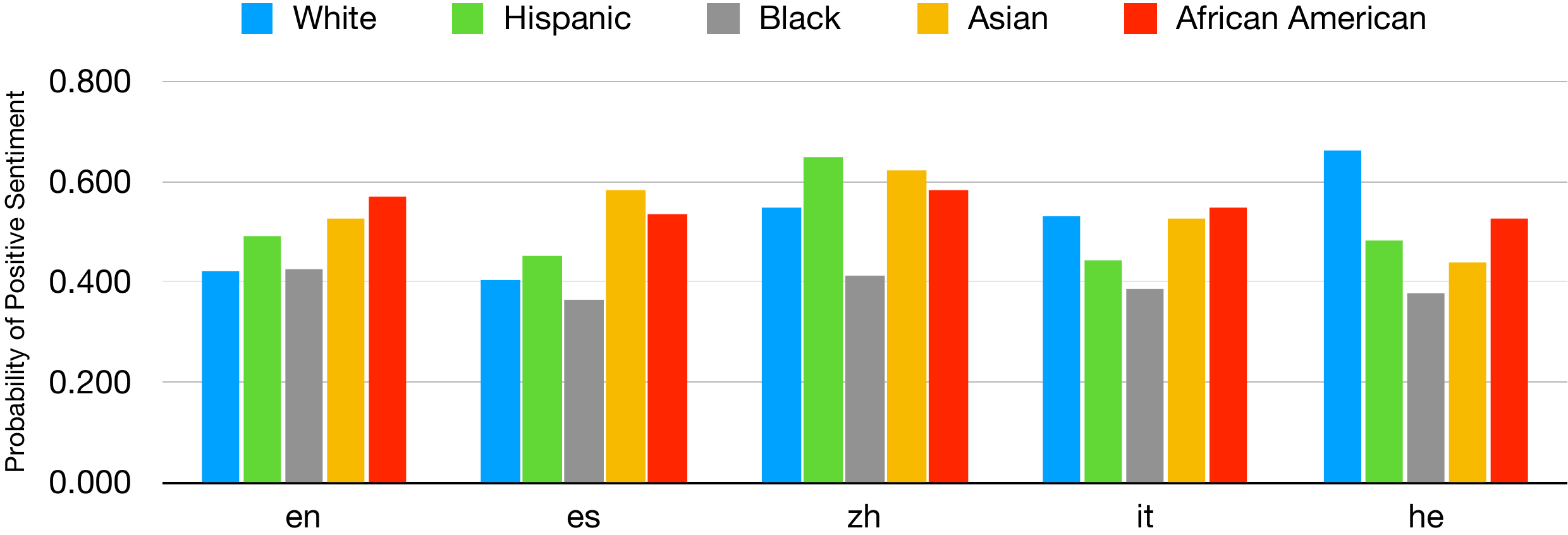}
\label{main:b}
\caption{Race}
\end{subfigure}

\caption{Religion (a) and race (b) predicted probabilities for female-subject templates with XLM-R. Language codes are English (en), Spanish (es), Chinese (zh), Italian (it), and Hebrew (he).}
\label{fig:phase2_example}
\end{figure*}

\subsection{Phase 1}
 Table \ref{tab:phase1_results} shows the result of finetuning mBERT and XLM-R for sentiment analysis. We use McNemar's test \cite{McNemar1947} to compare classification predictions between pairs of languages and find that English and Hebrew have significantly different (p~<~0.05) results for mBERT and XLM-R. In general, the resulting p-values between Hebrew and other languages are smaller than other pairs of languages. Following this, we create two sets of languages for our bias study in Phase 2: English, Chinese, Italian, and Spanish (Set 1) and Hebrew (Set 2). Set 1 consists of higher-performing languages and set 2 consists of a lower-performing language.
\subsection{Phase 2}
\paragraph{Overview}

To first determine whether there are any biases exhibited by these models for each language, we follow \citet{czarnowska-etal-2021-quantifying} and utilize Friedman's test to measure the statistical significance of the continuous probability scores across our groups for the race, religion, and nationality attributes. As gender contains only two groups, we use the Wilcox signed-rank test. These tests show that group probability scores for our models and languages are significantly different within race, religion, and nationality attributes but not gender. 

As we find a significant difference across group scores for three of our attributes, we are further interested in how different these scores are. Specifically, we study the variation of scores across groups for each attribute (MCM) to determine whether this variation affects some languages or attributes more than others. Our MCM scores, shown in Table \ref{tab:phase2_mcm}, provide evidence that \textbf{biases are not amplified or reduced more for a specific language, and instead, all languages are susceptible to exhibiting biases}.

In addition to measuring the amplification of existing biases, we investigate whether the biases are expressed similarly across the languages. To do so, we view whether groups favored/disfavored by one language are also favored/disfavored by other languages and in doing so, whether the group probability score distributions are similar. We show XLM-R probability score distributions (V) for female-subject religion and race templates in Figure \ref{fig:phase2_example} and those for nationality and gender in the Appendix. We find that there are some shared favored/disfavored groups. However, many biases are expressed differently for each language and model, such as atheists being favored in Spanish but not in English in XLM-R.

In the following text, we break down the results for each attribute and examine similarities across the high-performing languages (Set 1).

\paragraph{Race}
When examining our results for the race attribute, we are interested in whether any group is perceived more positively or negatively across all languages. Our VBCM results (Table \ref{tab:phase2_race}) show that \textbf{the 
Black race has more negative sentiment in comparison to the attribute average across all groups}. While we analyze African American as its own group within the race attribute, this can also be seen as a subset of Black. However, results for African American do not follow those of Black and are instead predicted more positively in comparison to other groups. As the two groups are not interchangeable, this may allude to differences in how the two groups are perceived.

\begin{table}
\centering
\small
\begin{tabular}{lcccc}
\hline
\textbf{Model} & \textbf{English} & \textbf{Spanish} & \textbf{Chinese} & \textbf{Italian} \\
\hline
mBERT(F) & -0.014  & -0.053  & -0.094 & -0.014 \\
mBERT(M) & -0.030 & -0.002 &  -0.088 & -0.038 \\

XLM-R(F) & -0.059  & -0.104  & -0.15 & -0.102 \\
XLM-R(M) &  -0.085 & -0.076 &  -0.145 & -0.078 \\
\hline
\end{tabular}
\caption{Black race VBCM results for female (F) and male (M) bias samples.}
\label{tab:phase2_race}
\end{table}

\paragraph{Religion}
We do not see any religions with consistently favored or disfavored groups across languages and models. However, we additionally use the MBCM metric to analyze the scores of majority religions against non-majority religions (Table \ref{tab:phase2_religion}). The results show that while the majority religions may be different across languages, \textbf{the sentiment for a language's majority religion is consistently more positive than for non-majority religions}.

\paragraph{Nationality}
Results for the nationality attribute show that \textbf{nationalities where the given language is an official language of the country have more positive sentiment in comparison to the attribute average} (VBCM). This follows our findings for religion, in which groups favored by a language are dominant within that language's culture. Countries within our nationality groups with an official language from our analysis list are: English (United States, Australia, India, Canada), Spanish (Mexico, Spain), Chinese (China), and Italian (Italy).

We show the detailed VBCM results in Table \ref{tab:phase2_nat} in the Appendix. While favored nationalities are not necessarily favored across all languages, all languages predict more positive sentiment for both American and Canadian nationalities in comparison to the attribute average. Meanwhile, the Russian nationality is not favored across languages and has more negative sentiment in comparison to the attribute average.

\begin{table}
\centering
\small
\begin{tabular}{lcccc}
\hline
\textbf{Model} & \textbf{English } & \textbf{Spanish} & \textbf{Chinese} & \textbf{Italian } \\
\hline
mBERT(F) & -0.081  & -0.043 & -0.009  & -0.083 \\
mBERT(M) & -0.112 & -0.039 & -0.013 & -0.088 \\

XLM-R(F) & -0.081  & -0.109  & -0.052  & -0.113 \\
XLM-R(M) & -0.074 &  -0.089 & -0.073 & -0.119 \\
\hline
\end{tabular}
\caption{Religion MBCM results for female (F) and male (M) bias samples.}
\label{tab:phase2_religion}
\end{table}

\paragraph{Gender}
When analyzing our gender-only bias samples\footnote{We align gender with biological sex in this setting.}, where we do not include mentions of race, religion, or nationality groups, we do not find any significant difference between male and female. However, this may be due to the small size of our gender-only samples, with 54 samples in total. 

As the other attributes contain parallel sets of samples with male and female subjects, we additionally analyze compound biases with gender and each of the three other attributes. Previous work has evaluated intersectional biases with gender as one component \cite{camara-etal-2022-mapping,honnavalli-etal-2022-towards}, though this has not been analyzed in non-English languages through explicit identity mentions.  We first analyze the probability distributions (V) for each attribute and compare the distributions with female versus male subjects. We find that the \textbf{distributions among groups are similar between the two genders} and groups favored by one gender are generally favored by the other.

While distributions among groups are similar, this does not indicate similar predicted probabilities. Our results show the \textbf{probabilities for paired samples of male and female subjects are significantly different} in many languages across race, religion, and nationality attributes, revealing a compound bias between gender and the other attributes. This occurs in both models, though XLM-R reduces this gender bias component in more cases.

\paragraph{Hebrew}
As Hebrew is a low-performing language for sentiment analysis in comparison to the other languages, we analyze it separately. Yet, we find that some results in Hebrew (Table \ref{tab:phase2_hebrew}) align with those observed across languages in Set 1. In particular, the attribute probability distributions are typically similar between genders but have significantly different predicted probabilities. Additionally, the Black race is also perceived more negatively and the only nationality with Hebrew as its official language, Israeli, is favored.

\subsection{Phase 3}
\paragraph{Finetuning}

When analyzing the effects of multilingual finetuning with MCM (Table \ref{tab:phase3_mcm}), we observe that biases across groups are amplified after multilingual finetuning for all languages and attributes (exception of race for English and Italian).

Viewing the predicted probabilities after monolingual and multilingual finetuning shows a positive skew after monolingual Chinese and Hebrew finetuning and multilingual finetuning on Hebrew samples (Figure \ref{fig:phase3_imbalance_a}). We hypothesize this is due to a label imbalance within the finetuning data, where there are more positive sentiment samples. To determine the impact of this label imbalance, we finetune additional models on label-balanced monolingual and multilingual data. We find that balancing the finetuning data can help mitigate this issue (Figure \ref{fig:phase3_imbalance_b}), and leads to average probabilities that are closer to the ground truth (0.5).

Our results also show that multilingual finetuning data significantly changes the group sentiment probabilities. Multilingual finetuning causes predicted probabilities to become more negative for all languages except Hebrew, which becomes more positive. As our multilingual finetuning data consists of data from several domains, we perform an additional experiment to determine the effect of multilingual finetuning on a single domain. Though we do not have sentiment analysis data from a single domain across all languages, MARC contains Amazon reviews in English, Spanish, and Chinese. We isolate these languages and show the results of finetuning on several versus one domain in Figure \ref{fig:phase3_domain}. We find that multilingual finetuning on a single domain diminishes the differences in probabilities.

\paragraph{Pretraining}
We analyze the two settings of multilingual pretraining: subsampled data and all data. The results of these pretraining settings in comparison to monolingual pretraining are shown in Table \ref{tab:phase3_pretraining} in the Appendix. We find that multilingual pretraining with subsampled data hurts classification across all languages. However, when pretraining on a combination of all monolingual finetuning data, classification performance exceeds that of monolingual pretraining for English.

\paragraph{Multilingual Data Impact}
When comparing the impacts of multilingual finetuning versus pretraining on biases observed during monolingual training, we notice greater effects with multilingual finetuning. Specifically, \textbf{multilingual finetuning amplifies bias in most cases while multilingual pretraining does not have a consistent outcome}.

We also find that \textbf{multilingual finetuning has a large impact on predicted sentiment probabilities}, where probabilities become more negative or positive in comparison to monolingual finetuning. Meanwhile, multilingual pretraining does not have this same effect and predicts probabilities similar to those with monolingual pretraining.

\section{Conclusion}
In this paper, we analyze various bias attributes across five languages on the downstream sentiment analysis task. Our results show that 1) bias is exhibited differently across different languages and models do not exhibit consistently low biases in a specific language and 2) models favor groups that are dominant within each language's culture. Together, these results provide evidence for the need to ground mitigation methods to specific languages/cultures instead of the findings from a single language. Further results show that multilingual finetuning data has more effect on bias amplification and changes in sentiment probabilities in comparison to multilingual pretraining data. We hope our findings encourage further diversity and expansion to additional languages in future bias studies. Additional future work can analyze culture-specific attributes (e.g. caste) and mitigate language-specific biases based on these results.

\section*{Limitations}
While we aim to provide an extensive study on biases across multiple languages, there are limitations to our work which we discuss below.

We describe our bias sample creation in Section \ref{sec:data}, where we detail our employment of human translators that are native speakers for each of our languages. We utilize one annotator per language and do not specify a required location for each annotator. As a result, the translations may be biased towards a particular localized variant of a language due to differences across regions. Expanding annotations to include several variants of each language can help us detect fine-grained biases within each language in our analysis.

While we analyze biases across several attributes and languages, our analysis can be improved through coverage of additional bias attributes such as sexual orientation and age. Additionally, we do not study attributes that may be specific to a subset of languages/cultures such as ethnic subgroups. Future work can expand on current attributes and examine the language-specific attributes in focused studies. We also limit our analysis to languages spoken natively by at least one of the authors. While this spans four language families (Germanic, Romance, Sino-Tibetan, Semitic), there are several language families that are not currently represented in our study.

To analyze biases, we adapt existing bias samples created in \citet{czarnowska-etal-2021-quantifying}. While they analyze genders beyond the binary male/female categories, we only consider these two genders in our analysis. As some of the languages we analyze are grammatically gendered, our usage of bias templates (described in Section \ref{sec:data}) will inherently describe subjects with either feminine or masculine nouns when translated from ambiguous language in English. As a result, comparisons of biases between a grammatically gendered language (Hebrew, Italian, Spanish) and languages that are not grammatically gendered (English, Chinese) will not be fair due to potential mismatching gender assumptions. We believe it is important to study gender bias beyond the binary gender as grammatically gendered languages are progressing towards the inclusion of other genders beyond masculine and feminine.

A final limitation we would like to discuss relates to the training data utilized in our study. While our test data and bias samples are parallel across languages, the data used to pretrain and finetune the models are not parallel and do not necessarily come from the same domains due to data and model availability. While differences in the data across these languages can inherently reflect varying societal biases for each language, the differences in domains can potentially amplify biases for certain languages due to discussions on specific topics.

\section*{Ethical Considerations}
To create our bias samples in the non-English languages, we source our annotations from a vendor that employs professional annotators native to each language. The vendor follows labor laws and employs annotators above the minimum wage. We separately validate the translations with internal researchers. We present the annotator instructions for translation in Figure \ref{fig:annotator_instructions}. The bias samples used in our study will be publicly released for future research studies under the Apache 2.0 license.

\bibliography{anthology, custom}
\bibliographystyle{acl_natbib}

\appendix
\section{Appendix}
\subsection{BLOOM Evaluation}
\paragraph{Prompting Details}
In addition to evaluating mBERT and XLM-R for biases in Phase 2, we also evaluate BLOOM\footnote{\url{https://huggingface.co/docs/transformers/model_doc/bloom}} on the subset of our languages included in the model's training: English, Chinese, and Spanish. In evaluating BLOOM, we aim to analyze biases in the multilingual auto-regressive setting with a large-scale model. To do so, we analyze the model's output during greedy decoding when predicting text among the three classes: positive, negative, and neutral. When evaluating BLOOM for bias, we use the following prompt in the zero-shot setting: ``Does the following text express positive, neutral, or negative sentiment?\escape{n}Text: \{Bias sample in any language\}\escape{n}The sentiment of the text is \{''. For all languages, the prompt is written in English, and the bias sample is written in one of the three languages. We include curly brackets around the bias sample and at the end of the prompt as previous work has shown the usage of curly brackets leads to higher quality generated text \cite{suzgun2022prompt}. Without the curly brackets, the generated text did not always fall into one of the three classes.

\paragraph{Results}
We show classification results with BLOOM in Table \ref{tab:phase2_bloom}. Although our prompt includes positive, negative, and neutral labels as options for classification, we find that the model rarely predicts neutral as the label. As a result, we include binary classification results as well. These results show a high accuracy for English bias samples, lower accuracy for Spanish, and random chance accuracy for Chinese samples. While we do not compare accuracy results among groups for each attribute, we find the model cannot distinguish sentiment varieties in Chinese and performs poorly in Spanish as well. Among the attributes, the model performs best on religion bias samples across all three languages. 

\begin{table}
\centering
\small
\begin{tabular}{lccc}
\hline
\textbf{Attribute} & \textbf{EN (3/2)} & \textbf{ES (3/2)} & \textbf{ZH (3/2)} \\
\hline
Gender & 0.537/0.805 & 0.407/0.611 & 0.333/0.500\\
Race & 0.485/0.727 & 0.485/0.727 & 0.333/0.500\\
Religion & 0.564/0.846 & 0.502/0.754 & 0.346/0.519\\
Nationality & 0.505/0.758 & 0.440/0.660 & 0.333/0.500 \\
\hline
\end{tabular}
\caption{3-way and 2-way classification accuracy results for English, Spanish, and Chinese with BLOOM.}
\label{tab:phase2_bloom}
\end{table}

\subsection{Data}\label{sec:appendix_data}
\paragraph{Fine-tuning Data}
The Multilingual Amazon Review Corpus \cite{keung-etal-2020-multilingual} is used to fine-tune in English, Chinese, and Spanish. The data consists of Amazon reviews and we assign reviews with a star rating of less than three as negative sentiment and those with more than three stars as positive sentiment. For Hebrew, we utilize a human-annotated dataset of comments on Facebook pages of political figures \cite{amram-etal-2018-representations}. These are already labeled with their corresponding sentiment (positive, negative, neutral). 

We utilize two sentiment analysis datasets for Italian as they are both smaller in scale. SENTIPOLC \cite{Barbieri2016OverviewOT} contains tweets labeled for subjectivity, polarity, and irony. We utilize tweets labeled as only overall positive or only overall negative for our positive and negative samples, respectively. ABSITA \cite{Basile2018OverviewOT} contains hotel reviews and is labeled for aspect-based sentiment analysis with aspects such as cleanliness and price. We select text with a majority of positive reviews across all aspects ( > 50\% positive aspect reviews) as positive sentiment samples and those with a majority of negative reviews as negative sentiment samples.

\paragraph{Parallel Test Data}
As the given labels in the XED dataset equate to different emotions, we categorize the subtitles for positive and negative sentiment according to the paper: anticipation, joy, and trust as positive, and anger, disgust, fear, and sadness as negative. 

\paragraph{Wikipedia Data}
For all languages except English, we used Wikipedia dumps from May 20th, 2022. For English, we used the Wikipedia dump from March 1st, 2022.

\paragraph{Group Selection} To select the groups for each attribute, we start with the list of groups in \cite{czarnowska-etal-2021-quantifying}. We translate the identity terms corresponding to the groups for each language. We compute the total frequency counts for each group in each language's respective Wikipedia. The groups are then ranked by frequency and the top overlapping most frequent groups are chosen for each attribute. By following this process, we select groups that are relevant to all languages in our analysis. To select a majority religion for each language as the background, we follow the same process and compute the most frequent religion for each language within its respective Wikipedia.

\subsection{Training Details}
The models used in Phases 1 and 2 are bert-base-multilingual-cased (110 million parameters) and xlm-roberta-base (125 million parameters) from Hugging Face \cite{wolf-etal-2020-transformers}. When finetuning the models in Phase 1, we use a learning rate of 2e-5, weight decay of 0.01, finetune for 10 epochs, and save the model with the best accuracy on the validation set.

During Phase 3, we use bert-base-multilingual-cased and additionally pretrain our own BERT models (110 million parameters). The vocabulary size for our pretrained monolingual BERT is 30522, and we pretrain the monolingual Chinese BERT model with a limited alphabet of 20000. Our pretrained multilingual BERT model has a vocab size of 121806. The models are pretrained for 5 epochs with a batch size of 64, 1000 warmup steps, learning rate of 1e-4, and weight decay of 0.01.

\subsection{Additional Results}
\begin{figure*}

\begin{minipage}{.5\linewidth}
\centering
\subfloat[]{\label{main:c}\includegraphics[scale=.31]{images/phase2_xlmr_rel_female_2.pdf}}
\end{minipage}%
\begin{minipage}{.5\linewidth}
\centering
\subfloat[]{\label{main:d}\includegraphics[scale=.31]{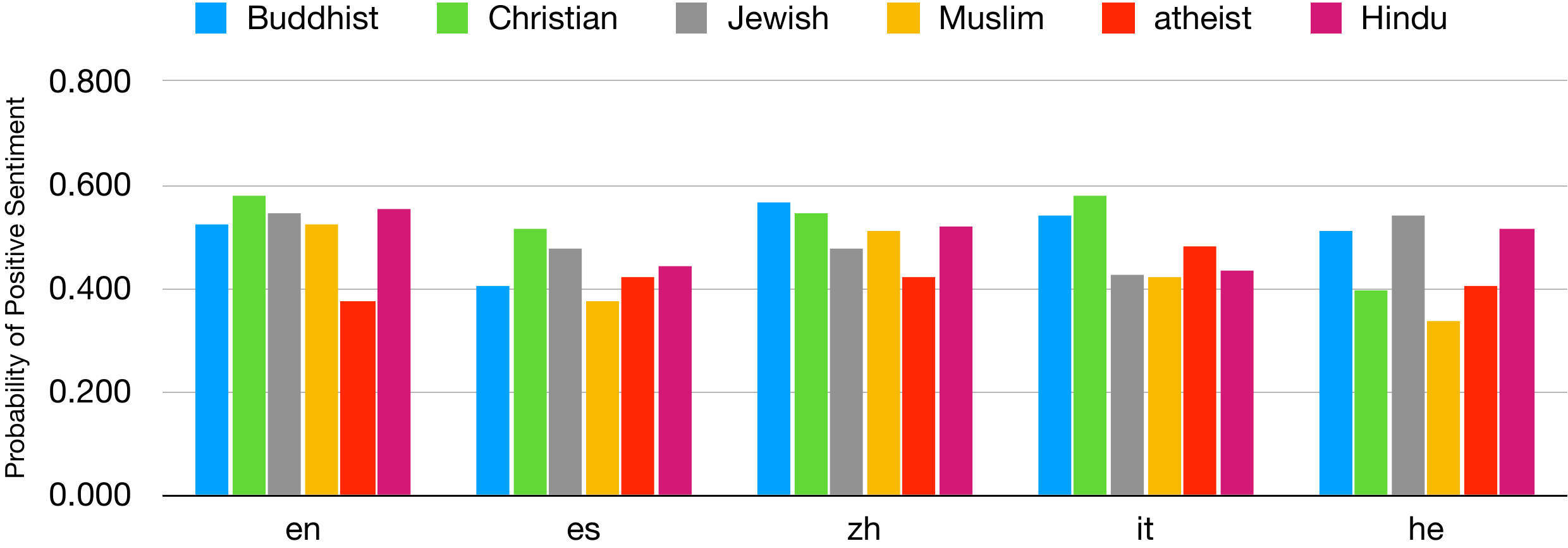}}
\end{minipage}\par\medskip

\begin{minipage}{.5\linewidth}
\centering
\subfloat[]{\label{main:a}\includegraphics[scale=.31]{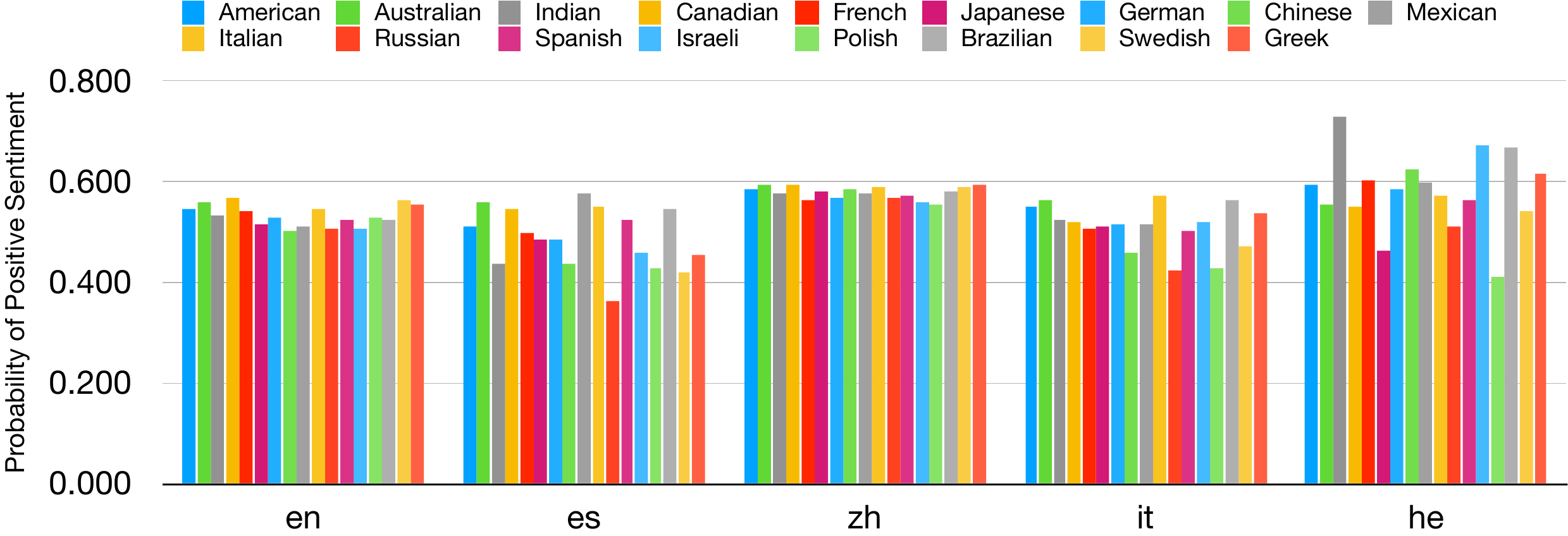}}
\end{minipage}%
\begin{minipage}{.5\linewidth}
\centering
\subfloat[]{\label{main:b}\includegraphics[scale=.31]{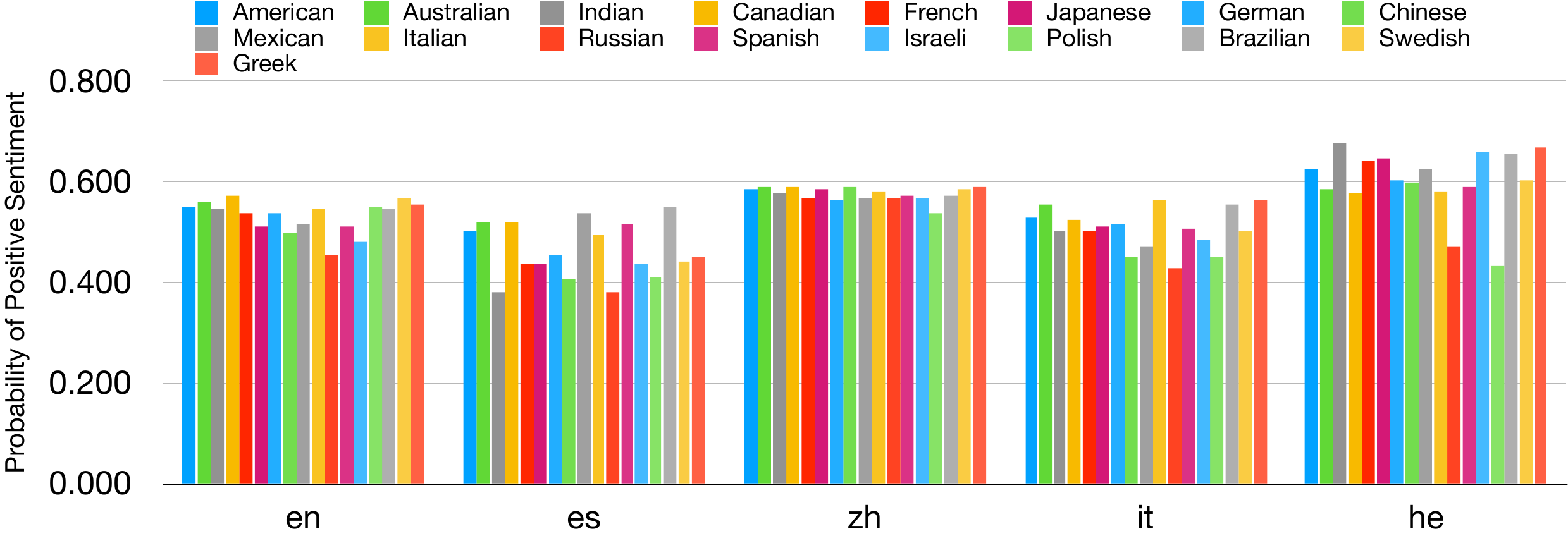}}
\end{minipage}\par\medskip

\begin{minipage}{.5\linewidth}
\centering
\subfloat[]{\label{main:c}\includegraphics[scale=.31]{images/phase2_xlmr_race_female.pdf}}
\end{minipage}%
\begin{minipage}{.5\linewidth}
\centering
\subfloat[]{\label{main:d}\includegraphics[scale=.31]{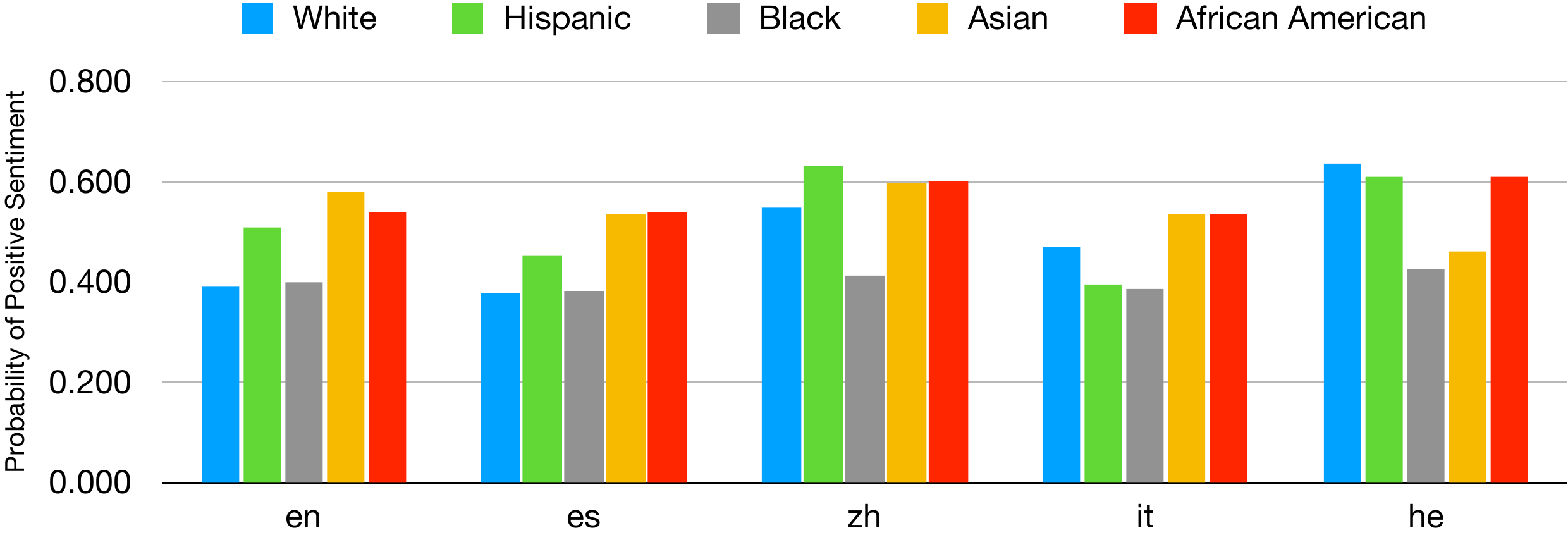}}
\end{minipage}\par\medskip

\begin{minipage}{\linewidth}
\centering
\subfloat[]{\label{main:c}\includegraphics[scale=.31]{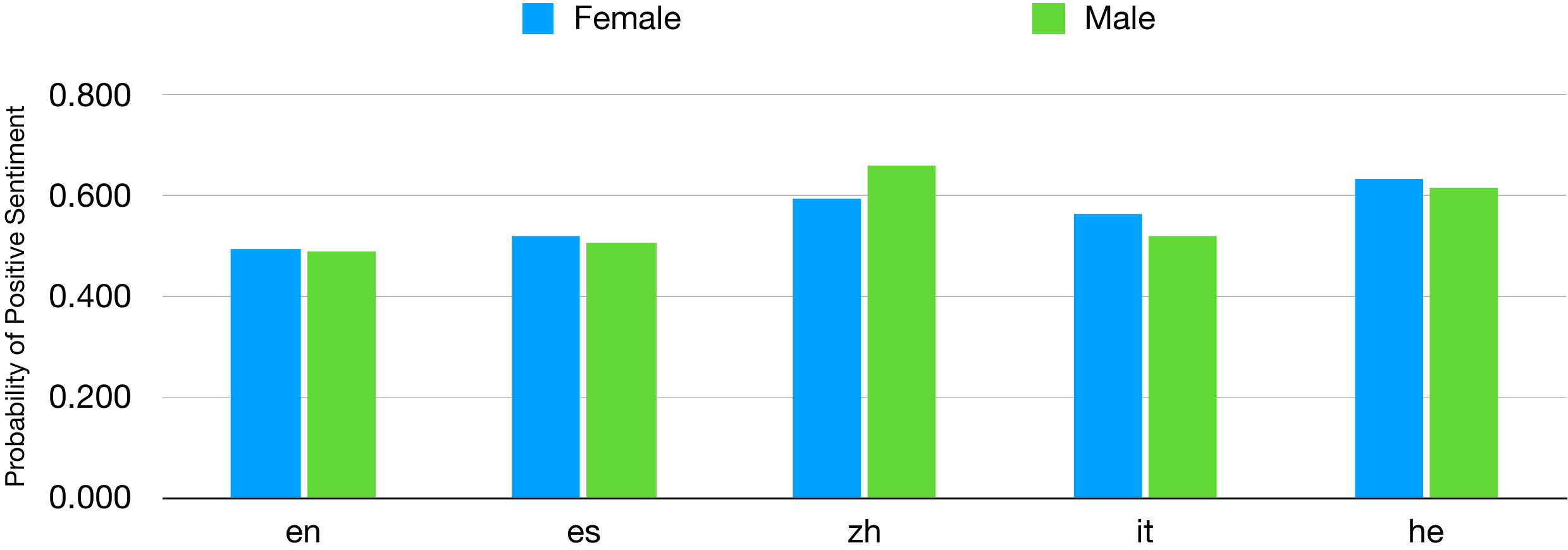}}
\end{minipage}

\caption{Phase 2 XLM-R results for female (left) and male (right) subjects.}
\label{fig:xlmr_all}
\end{figure*}

\begin{figure*}

\begin{minipage}{.5\linewidth}
\centering
\subfloat[]{\label{main:c}\includegraphics[scale=.31]{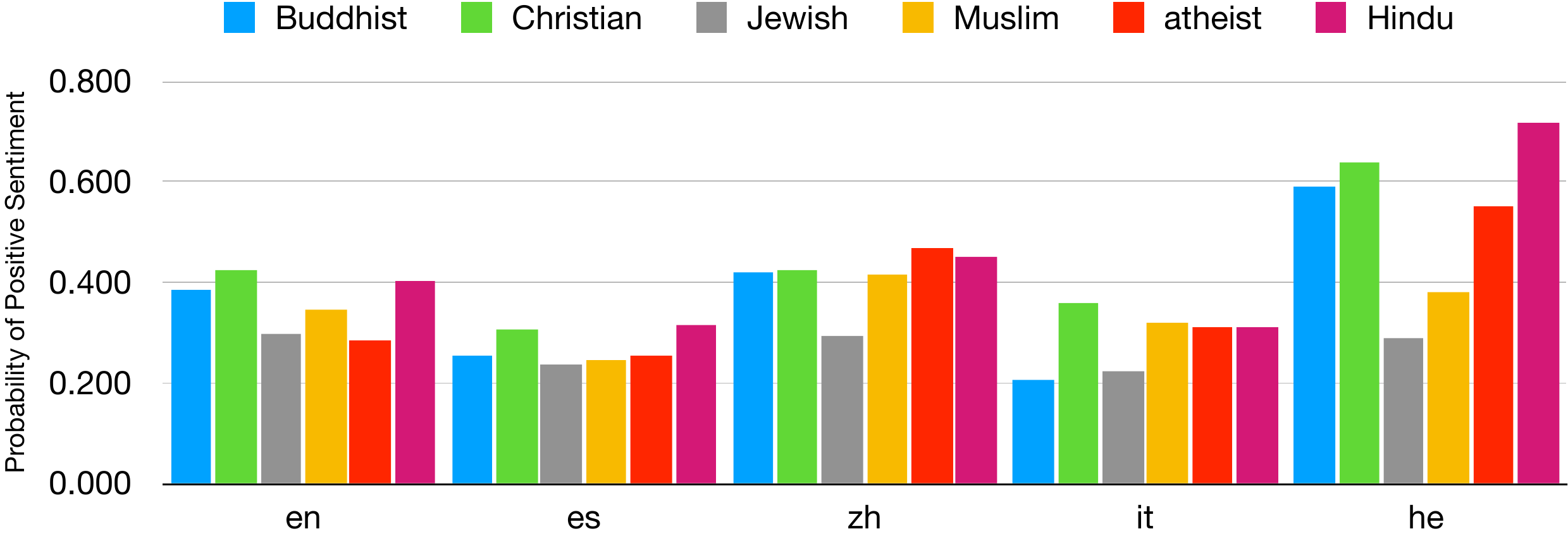}}
\end{minipage}%
\begin{minipage}{.5\linewidth}
\centering
\subfloat[]{\label{main:d}\includegraphics[scale=.31]{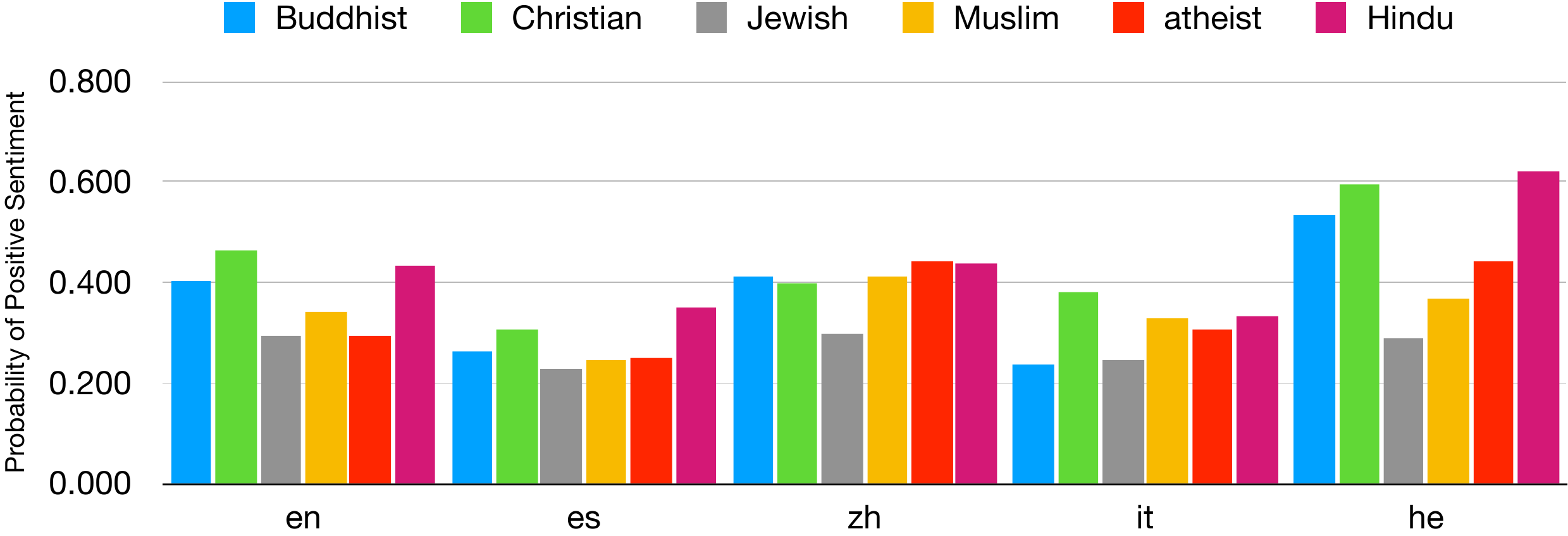}}
\end{minipage}\par\medskip

\begin{minipage}{.5\linewidth}
\centering
\subfloat[]{\label{main:a}\includegraphics[scale=.31]{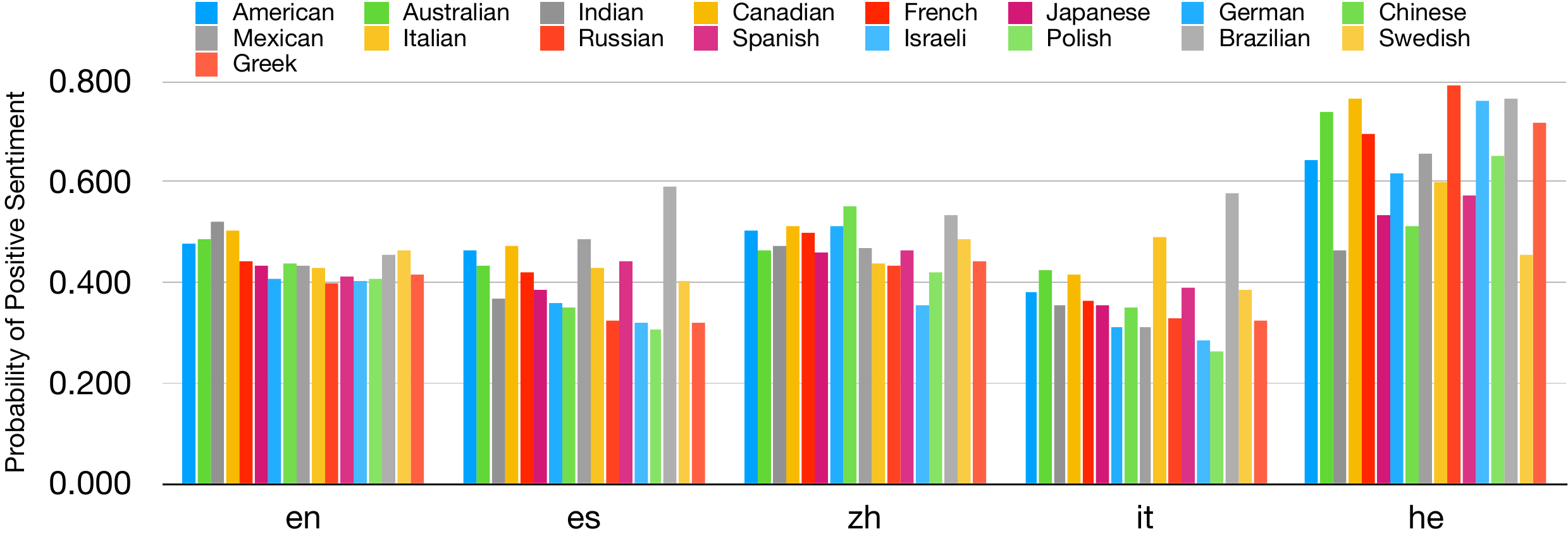}}
\end{minipage}%
\begin{minipage}{.5\linewidth}
\centering
\subfloat[]{\label{main:b}\includegraphics[scale=.31]{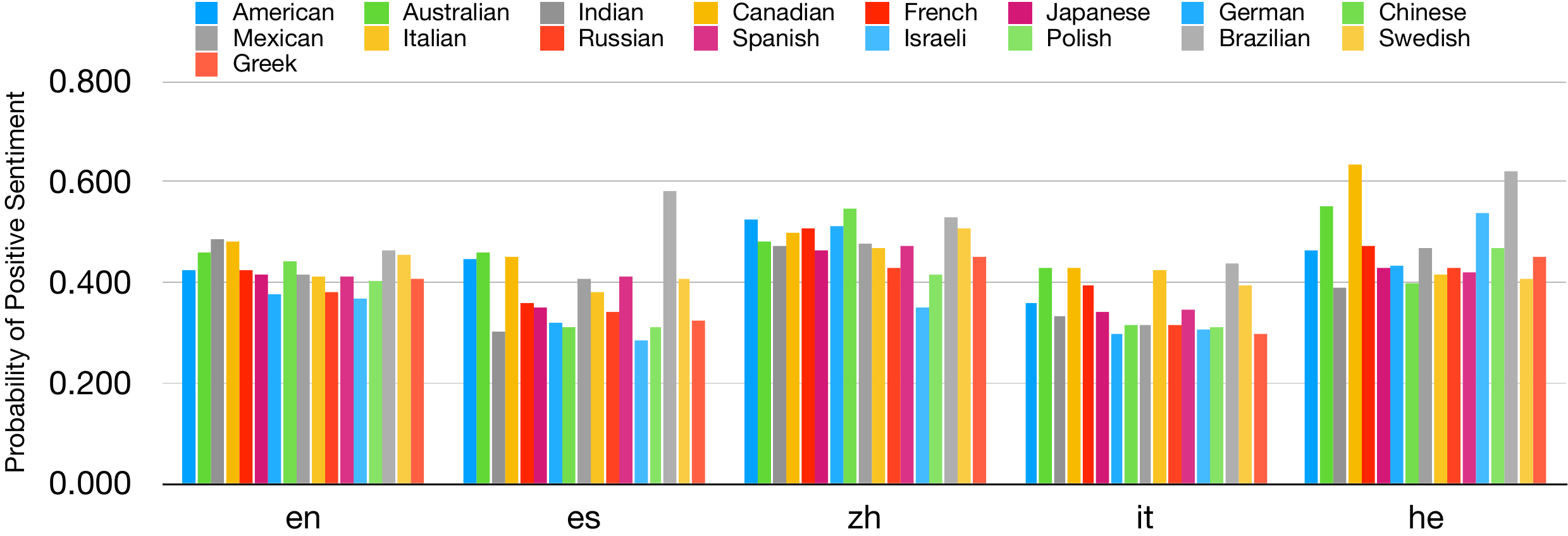}}
\end{minipage}\par\medskip

\begin{minipage}{.5\linewidth}
\centering
\subfloat[]{\label{main:c}\includegraphics[scale=.31]{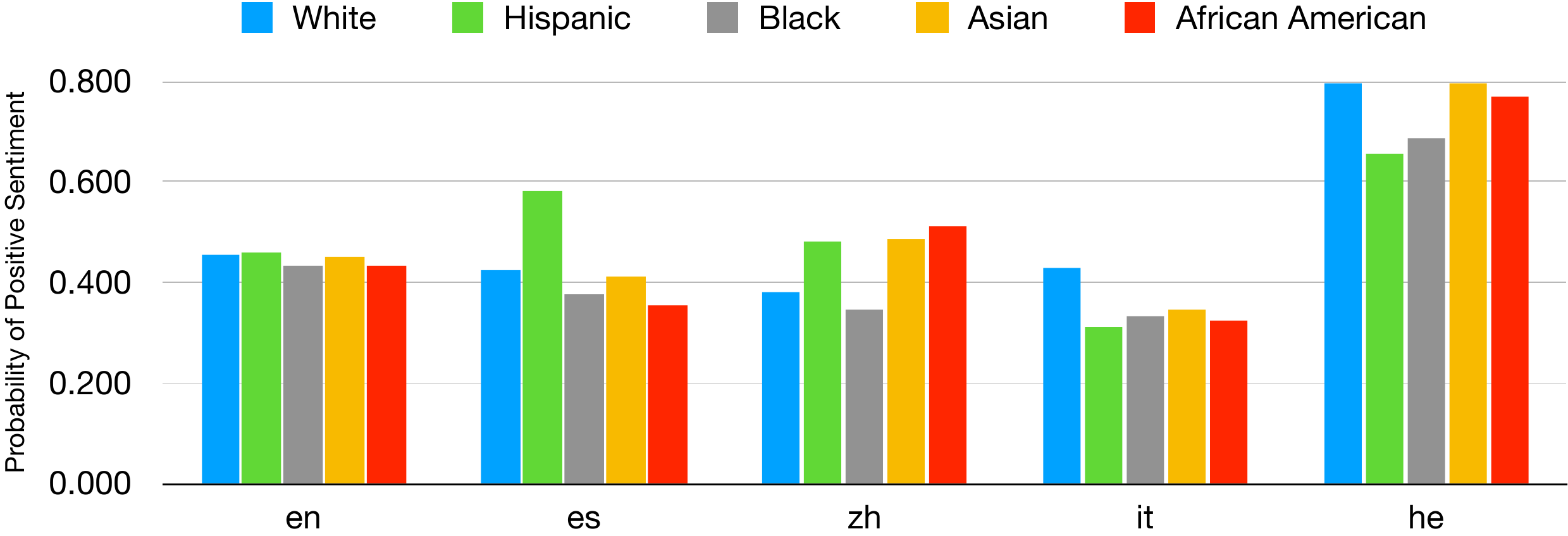}}
\end{minipage}%
\begin{minipage}{.5\linewidth}
\centering
\subfloat[]{\label{main:d}\includegraphics[scale=.31]{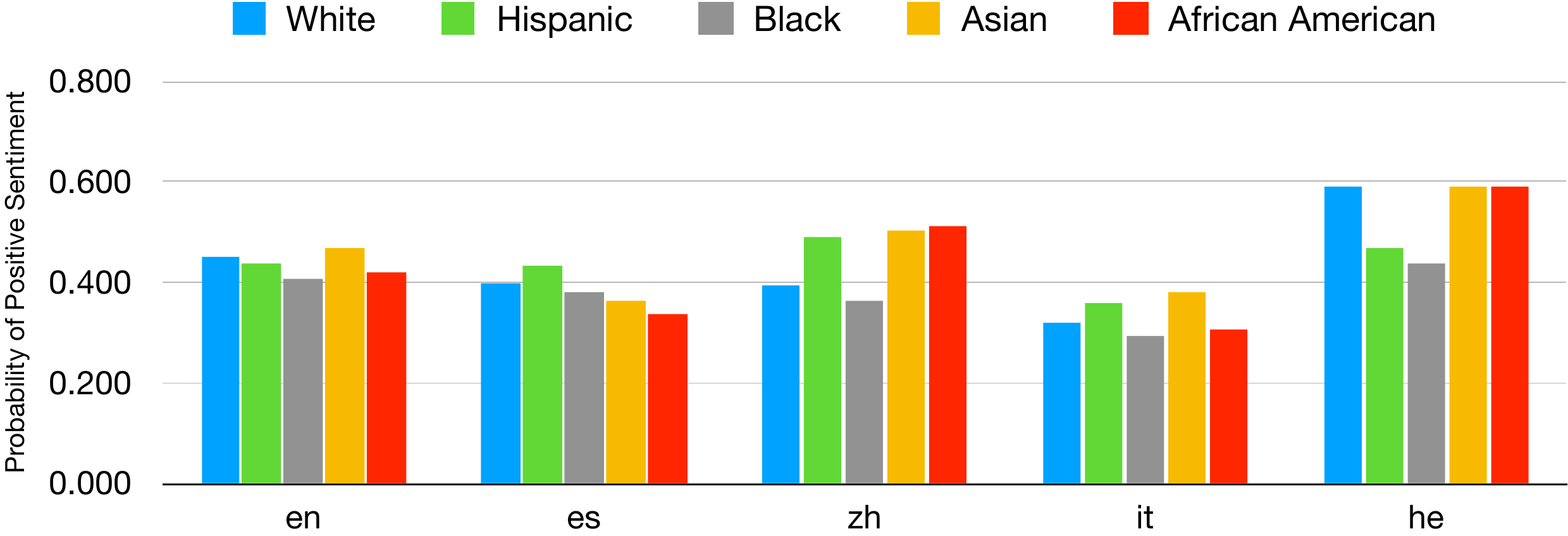}}
\end{minipage}\par\medskip

\begin{minipage}{\linewidth}
\centering
\subfloat[]{\label{main:c}\includegraphics[scale=.31]{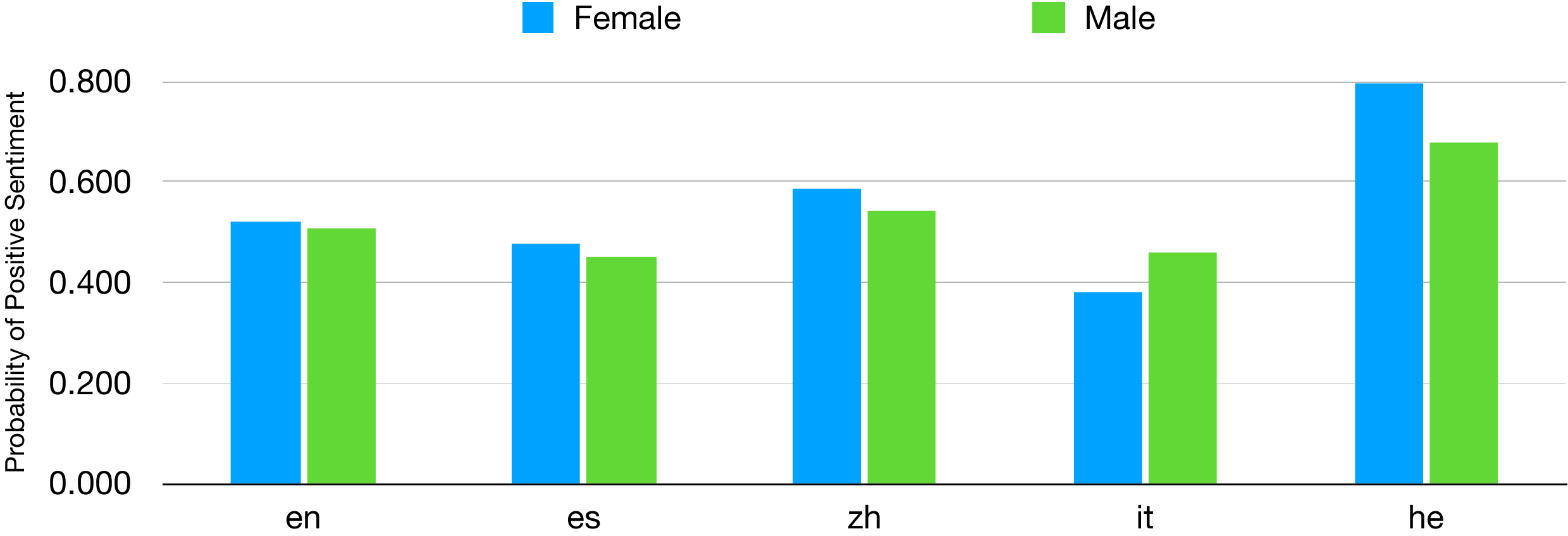}}
\end{minipage}

\caption{Phase 2 mBERT results for female (left) and male (right) subjects.}
\label{fig:mbert_all}
\end{figure*}

\begin{figure*}

\begin{minipage}{.5\linewidth}
\centering
\subfloat[]{\label{main:c}\includegraphics[scale=.31]{images/phase3_finetune_race_female.pdf}}
\end{minipage}%
\begin{minipage}{.5\linewidth}
\centering
\subfloat[]{\label{main:d}\includegraphics[scale=.31]{images/phase3_finetune_equal_race_female.pdf}}
\end{minipage}\par\medskip

\begin{minipage}{.5\linewidth}
\centering
\subfloat[]{\label{main:a}\includegraphics[scale=.31]{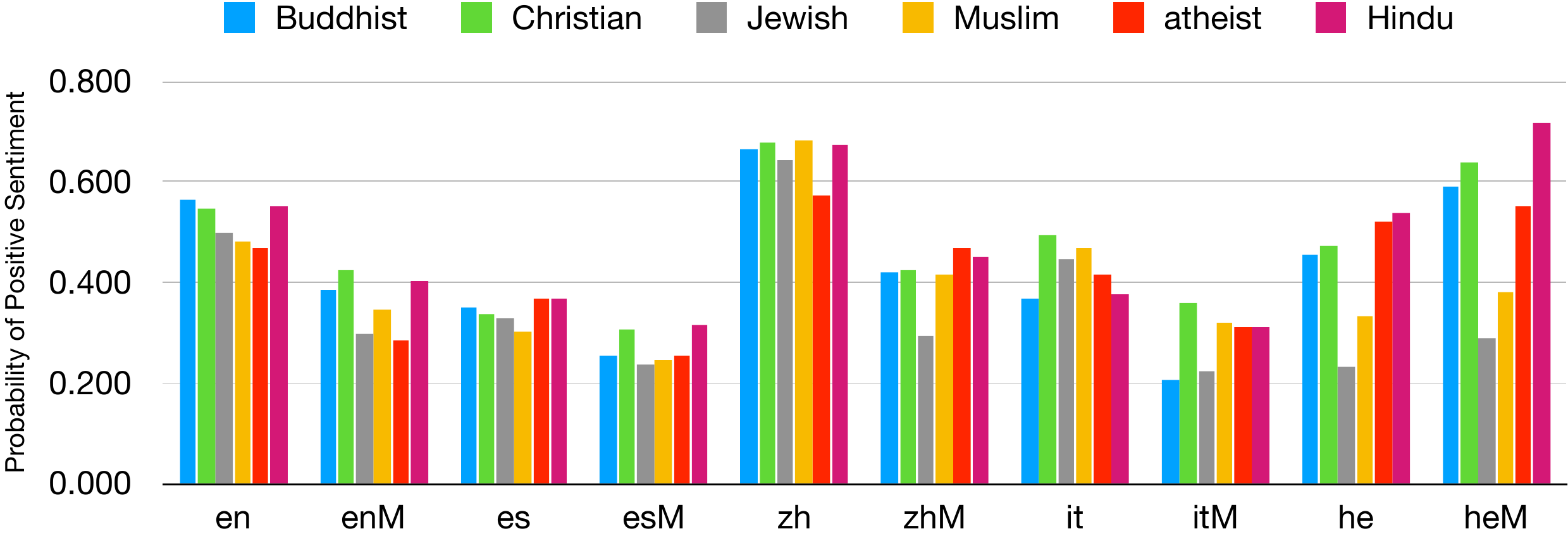}}
\end{minipage}%
\begin{minipage}{.5\linewidth}
\centering
\subfloat[]{\label{main:b}\includegraphics[scale=.31]{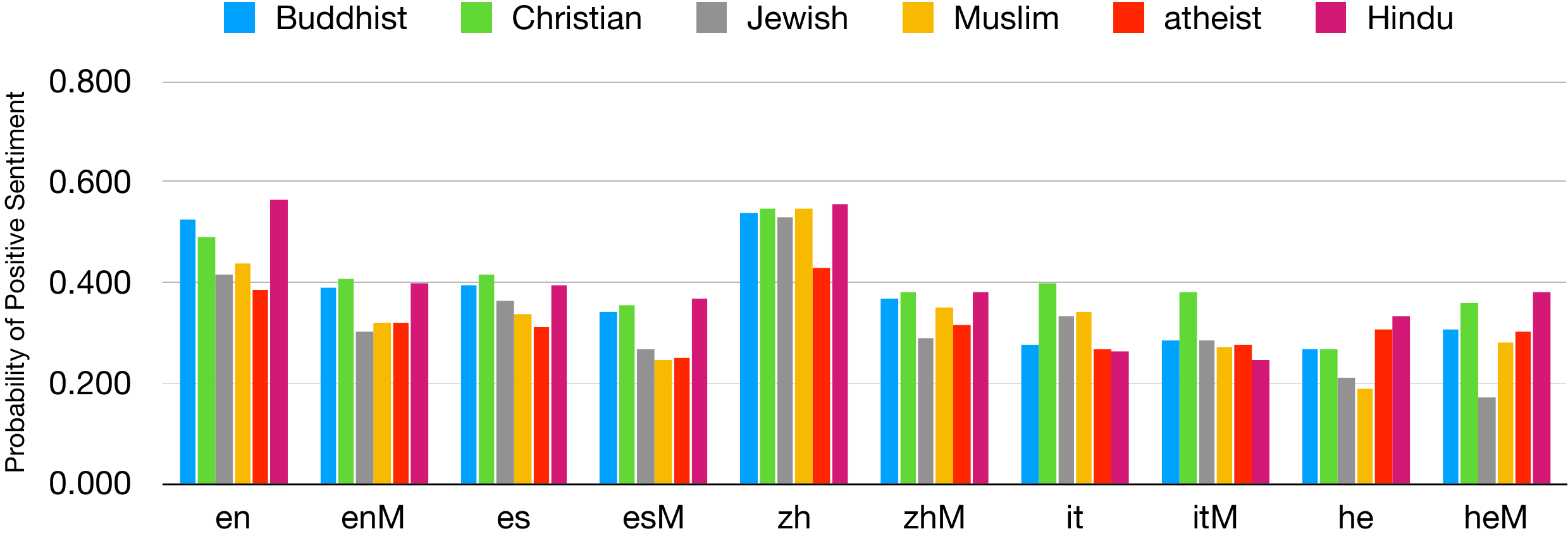}}
\end{minipage}\par\medskip

\begin{minipage}{.5\linewidth}
\centering
\subfloat[]{\label{main:c}\includegraphics[scale=.31]{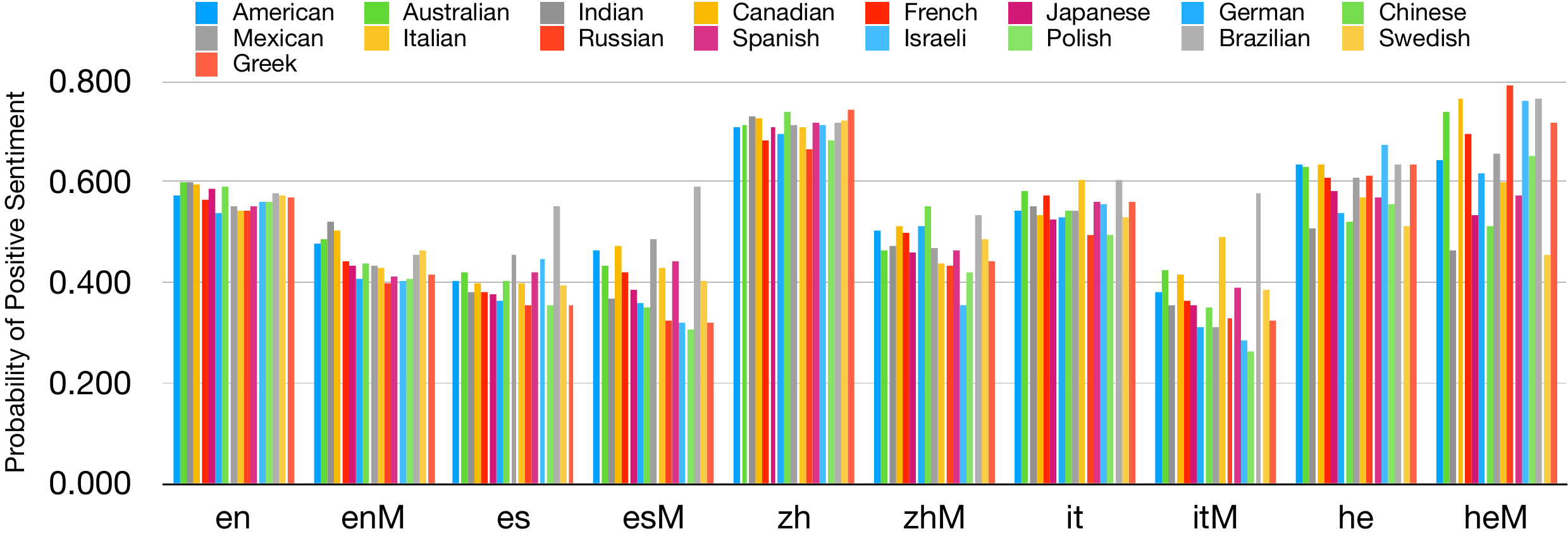}}
\end{minipage}%
\begin{minipage}{.5\linewidth}
\centering
\subfloat[]{\label{main:d}\includegraphics[scale=.31]{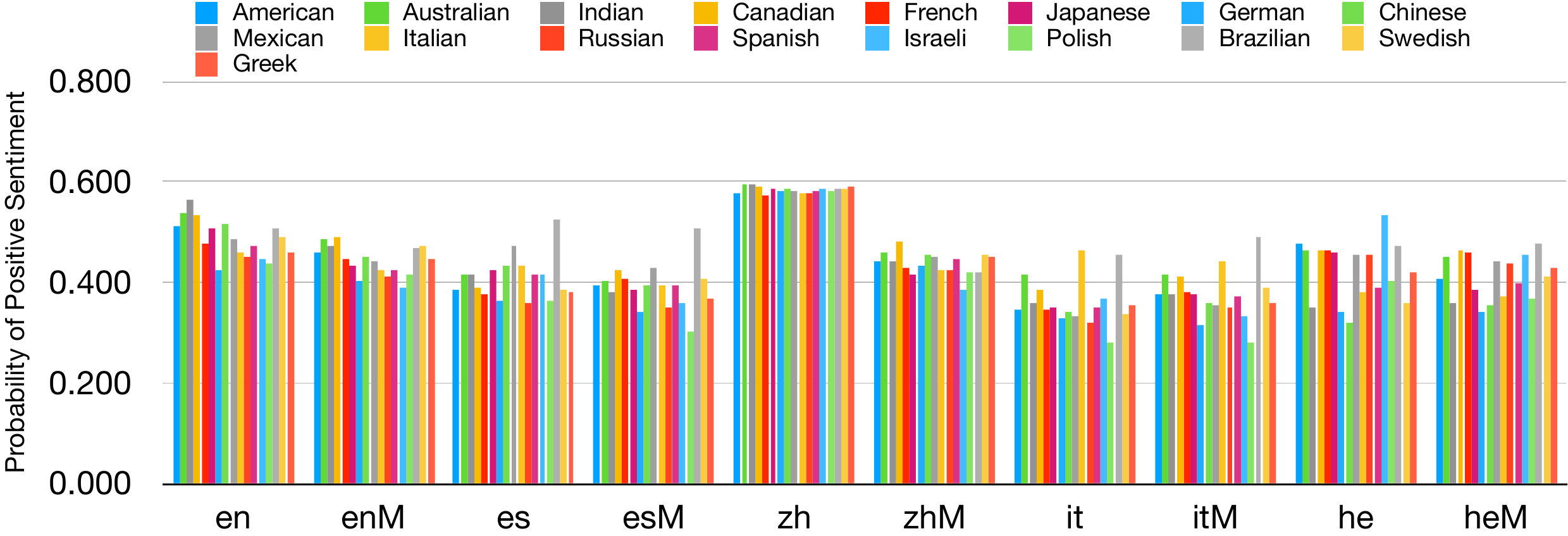}}
\end{minipage}\par\medskip

\begin{minipage}{.5\linewidth}
\centering
\subfloat[]{\label{main:c}\includegraphics[scale=.31]{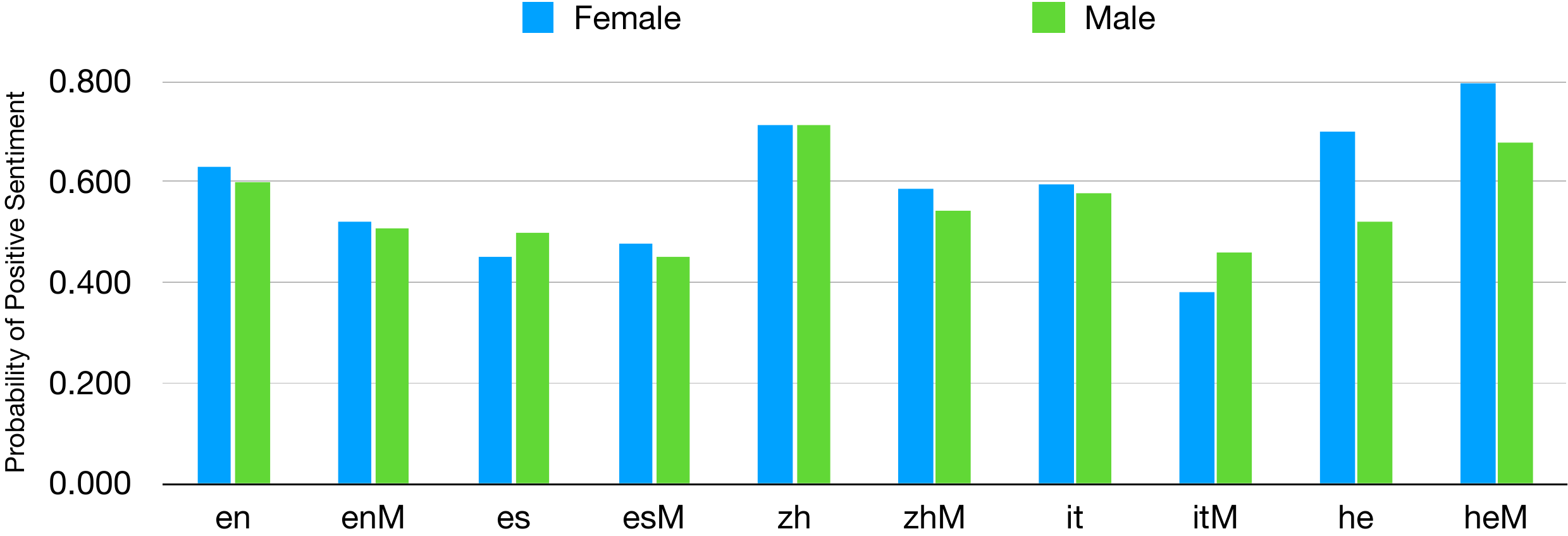}}
\end{minipage}%
\begin{minipage}{.5\linewidth}
\centering
\subfloat[]{\label{main:d}\includegraphics[scale=.31]{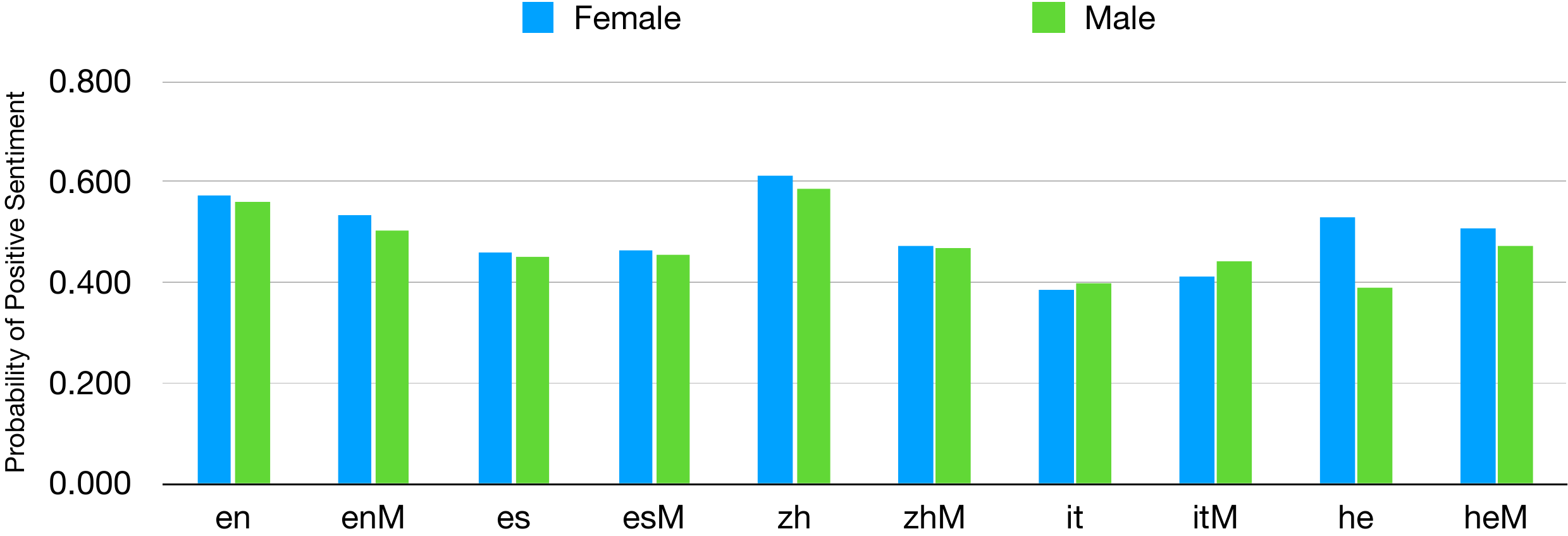}}
\end{minipage}\par\medskip

\caption{Predicted probabilities before (left) and after (right) label balancing for female-subject templates.}
\label{fig:balanced_all}
\end{figure*}

\begin{figure*}

\begin{minipage}{.5\linewidth}
\centering
\subfloat[]{\label{main:c}\includegraphics[scale=.31]{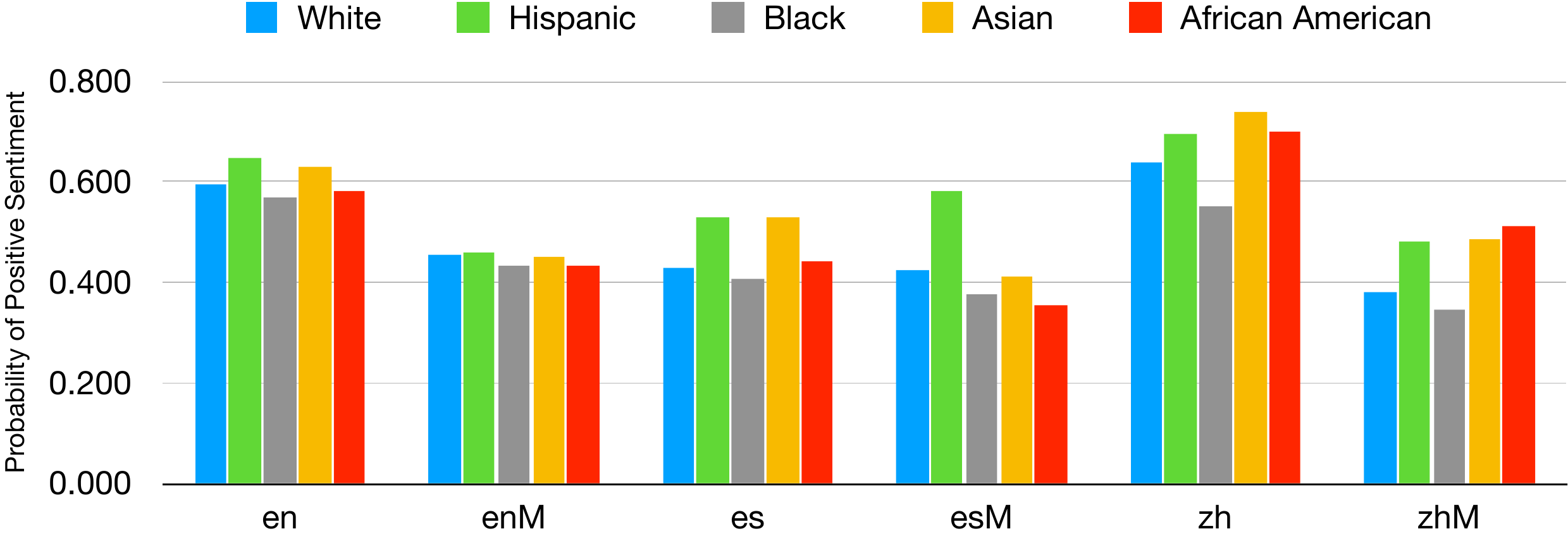}}
\end{minipage}%
\begin{minipage}{.5\linewidth}
\centering
\subfloat[]{\label{main:d}\includegraphics[scale=.31]{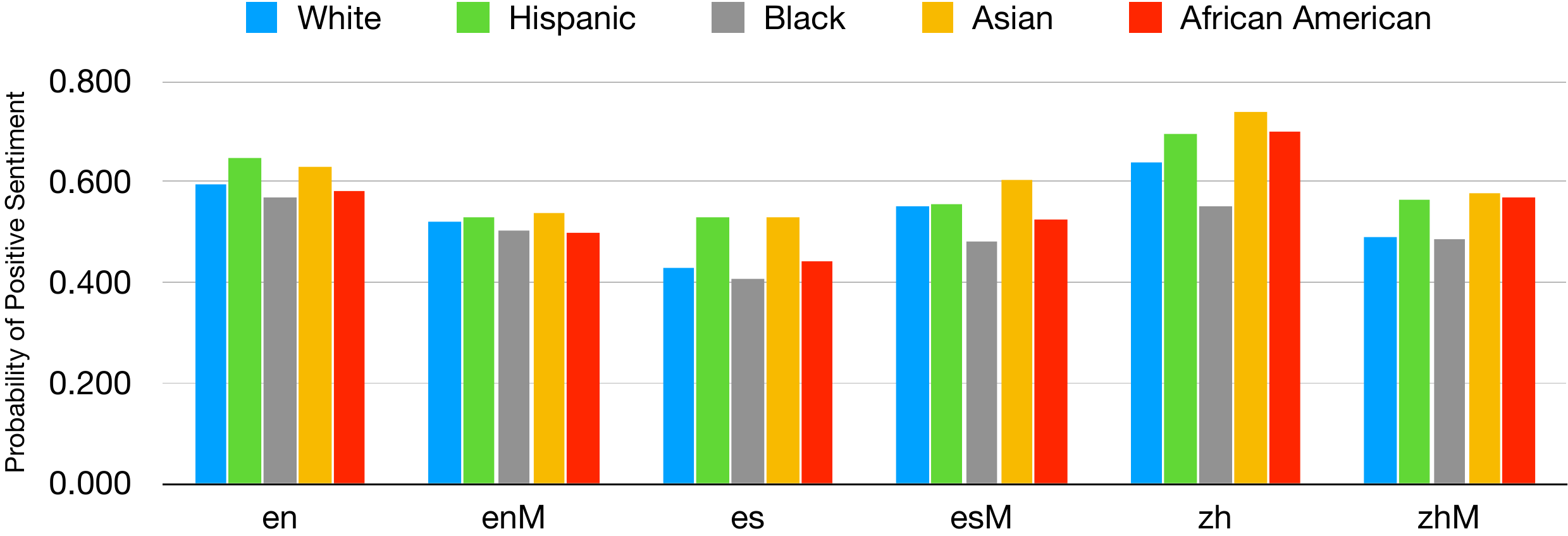}}
\end{minipage}\par\medskip

\begin{minipage}{.5\linewidth}
\centering
\subfloat[]{\label{main:c}\includegraphics[scale=.31]{images/phase3_finetune_multidomain_rel_female.pdf}}
\end{minipage}%
\begin{minipage}{.5\linewidth}
\centering
\subfloat[]{\label{main:d}\includegraphics[scale=.31]{images/phase3_finetune_single_rel_female.pdf}}
\end{minipage}\par\medskip

\begin{minipage}{.5\linewidth}
\centering
\subfloat[]{\label{main:c}\includegraphics[scale=.31]{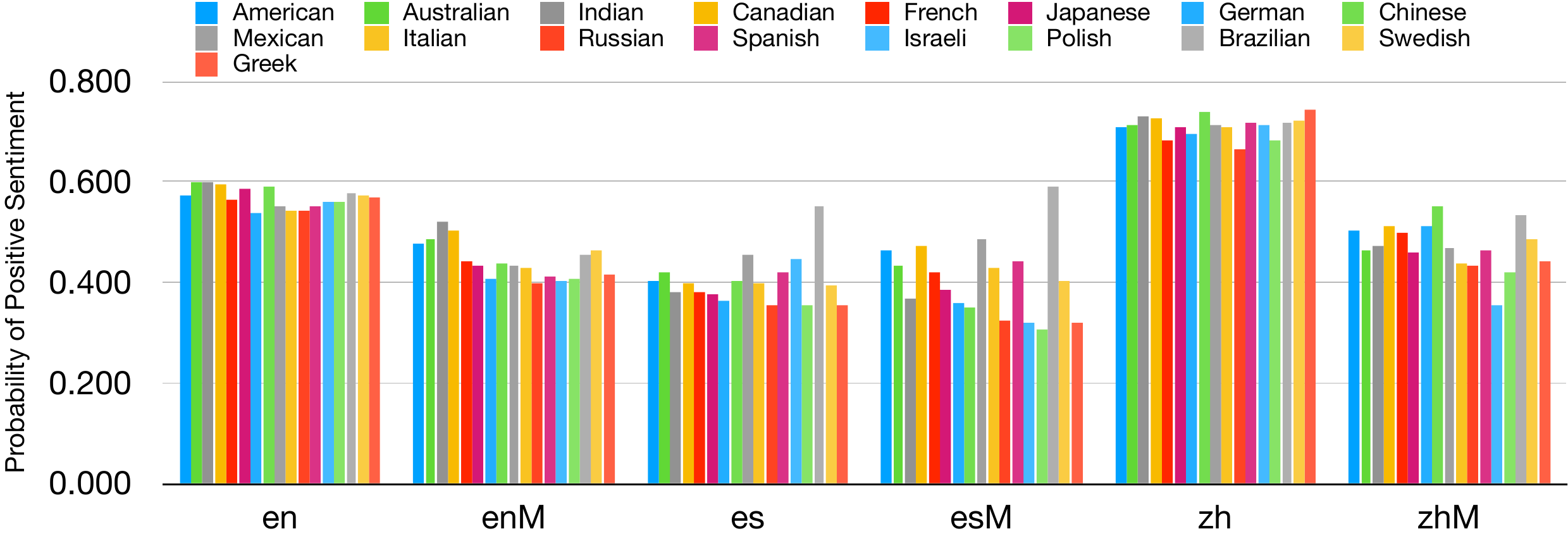}}
\end{minipage}%
\begin{minipage}{.5\linewidth}
\centering
\subfloat[]{\label{main:d}\includegraphics[scale=.31]{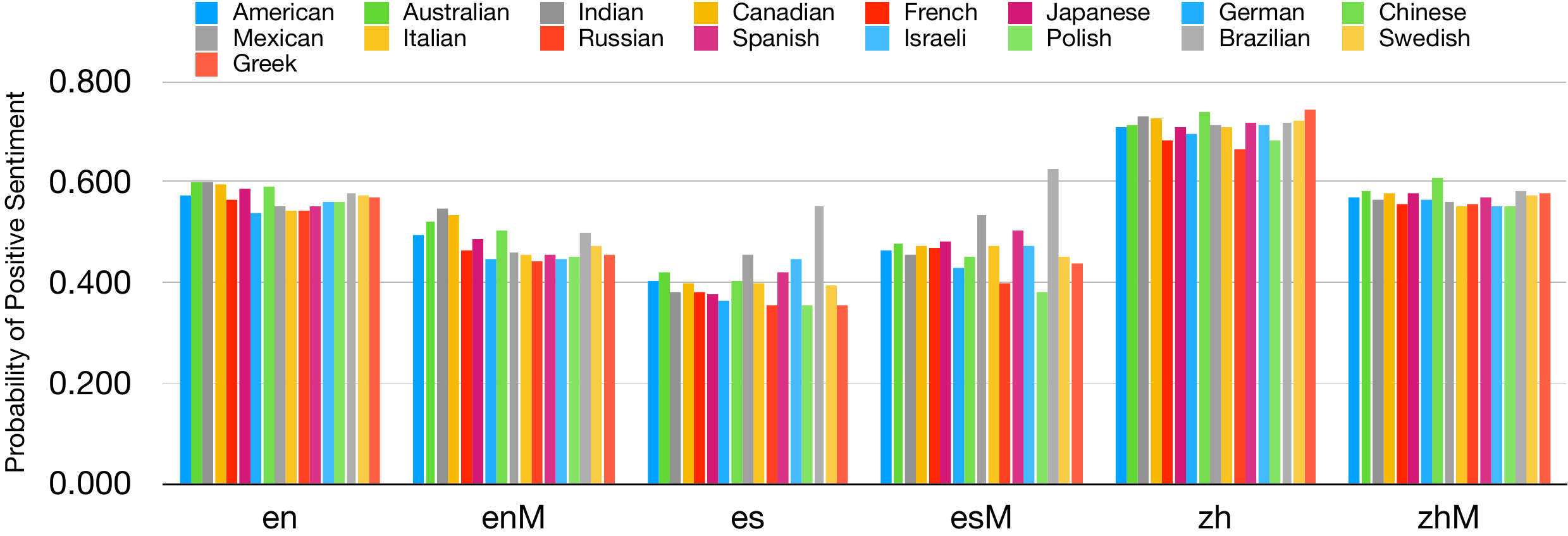}}
\end{minipage}\par\medskip

\begin{minipage}{.5\linewidth}
\centering
\subfloat[]{\label{main:c}\includegraphics[scale=.31]{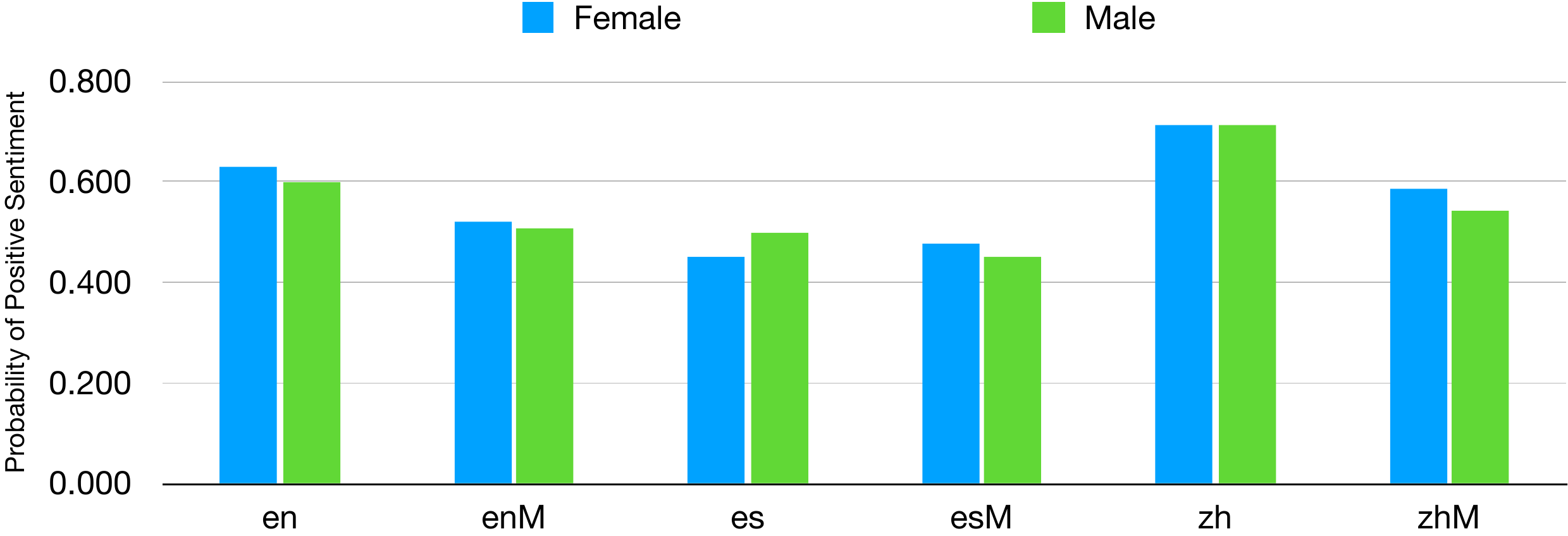}}
\end{minipage}%
\begin{minipage}{.5\linewidth}
\centering
\subfloat[]{\label{main:d}\includegraphics[scale=.31]{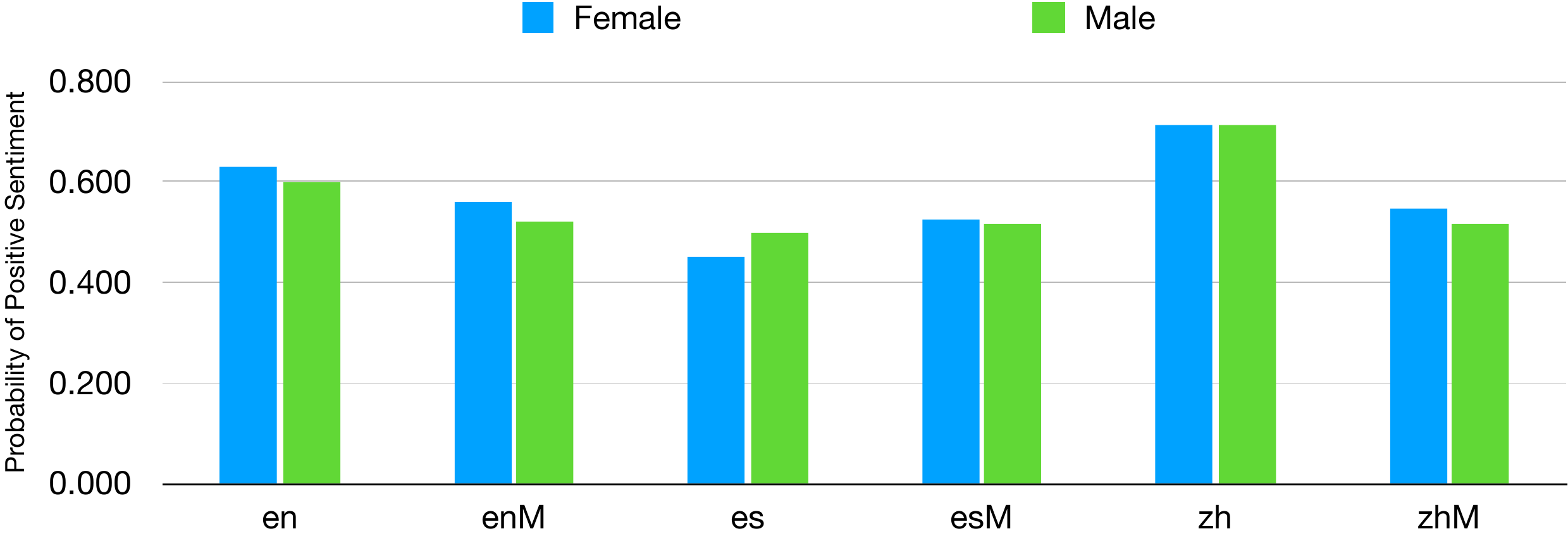}}
\end{minipage}\par\medskip

\caption{Predicted probabilities after finetuning on multiple domains (left)  and a single domain (right).}
\label{fig:domain_all}
\end{figure*}

We show male/female-subject results for all attributes in Phases 2 and 3 in Figures \ref{fig:xlmr_all}, \ref{fig:mbert_all}, \ref{fig:balanced_all}, and \ref{fig:domain_all}.

\begin{table}
\centering
\small
\begin{tabular}{lcc}
\hline
\textbf{Split} & \textbf{Sentiment} & \textbf{\# Samples}\\
\hline
Train & positive & 4425 \\
& negative & 2193 \\
Validation & positive & 1700 \\
& negative & 612 \\
Test & positive & 79 \\
& negative & 108 \\
\hline
\end{tabular}
\caption{Dataset statistics for fine-tuning train, validation, and test sets. Training and validation data come from various domains while the test set is parallel across languages.}
\label{tab:datasets}
\end{table}

\begin{table}
\centering
\small
\begin{tabular}{lcc}
\hline
\textbf{Language} & \textbf{mBERT} & \textbf{XLM-R}\\
\hline
English & 0.787 & 0.817 \\
Hebrew & 0.634 & 0.691 \\
Chinese & 0.726 & 0.806 \\
Italian & 0.734 & 0.759 \\
Spanish & 0.717 & 0.808 \\
\hline
\end{tabular}
\caption{Accuracy results on a parallel test set after finetuning mBERT and XLM-R on multilingual sentiment analysis data. Results are averaged over 3 finetuned models for mBERT and XLM-R.}
\label{tab:phase1_results}
\end{table}

\begin{table}[t!]
\centering
\small
\begin{tabular}{llc}
\hline
\textbf{Group} & \textbf{Model} & \textbf{Hebrew (F/M)} \\
\hline
Black & mBERT & -0.062 / -0.098 \\
& XLM-R & -0.120 / -0.122 \\
\hline
Israeli & mBERT & 0.115 / 0.066 \\
& XLM-R & 0.091 / 0.059 \\
\hline
\end{tabular}
\caption{VBCM probability results for Black race and Israeli nationality bias samples in Hebrew.}
\label{tab:phase2_hebrew}
\end{table}

\begin{table*}
\centering
\small
\begin{tabular}{llccccc}
\hline

\textbf{Nationality} & \textbf{Model} & \textbf{English (F/M)} & \textbf{Spanish (F/M)} & \textbf{Chinese (F/M)} & \textbf{Italian (F/M)}  \\
\hline
American & mBERT & 0.032 / 0.001 & 0.06 / 0.068 & 0.031 / 0.05 & 0.011 / 0.002 \\

& XLM-R & 0.012 / 0.018 & 0.025 / 0.041 & 0.007 / 0.007 & 0.039 / 0.024 \\

Australian & mBERT & 0.043 / 0.033 & 0.029 / 0.081 & -0.006 / 0.003 & 0.054 / 0.074 \\

& XLM-R & 0.026 / 0.027 & 0.073 / 0.056 & 0.018 / 0.015 & 0.053 / 0.049 \\

Indian & mBERT & 0.077 / 0.063 & -0.039 / -0.079 & 0.003 / -0.003 & -0.017 / -0.024 \\

& XLM-R & 0.001 / 0.013 & -0.051 / -0.084 & 0.001 / 0.002 & 0.014 / -0.005 \\

Canadian & mBERT & 0.063 / 0.055 & 0.07 / 0.073 & 0.039 / 0.02 & 0.047 / 0.075 \\

& XLM-R & 0.035 / 0.040 & 0.058 / 0.057 & 0.014 / 0.014 & 0.007 / 0.019 \\
\hline
Mexican & mBERT & -0.01 / -0.008 & 0.081 / 0.028 & -0.001 / -0.001 & -0.060 / -0.040 \\

& XLM-R & -0.022 / -0.015 & 0.091 / 0.073 & -0.003 / -0.009 & 0.004 / -0.036 \\

Spanish & mBERT & -0.031 / -0.013 & 0.039 / 0.031 & -0.009 / -0.005 & 0.017 / -0.010 \\

& XLM-R & -0.010 / -0.021 & 0.036 / 0.051 & -0.006 / -0.004 & -0.010 / 0.000 \\
\hline
Chinese & mBERT & -0.005 / 0.017 & -0.054 / -0.07 & 0.078 / 0.069 & -0.021 / -0.041 \\

& XLM-R & -0.029 / -0.035 & -0.05 / -0.055 & 0.006 / 0.014 & -0.051 / -0.055 \\
\hline
Italian & mBERT & -0.013 / -0.015 & 0.023 / 0.000 & -0.034 / -0.009 & 0.118 / 0.070 \\

& XLM-R & 0.014 / 0.016 & 0.062 / 0.032 & 0.010 / 0.006 & 0.060 / 0.057 \\
\hline
Russian & mBERT & -0.045 / -0.046 & -0.079 / -0.039 & -0.039 / -0.047 & -0.042 / -0.04 \\

& XLM-R & -0.026 / -0.076 & -0.124 / -0.084 & -0.01 / -0.01 & -0.088 / -0.08 \\
\hline
\end{tabular}
\caption{VBCM probability results for nationality bias samples.}
\label{tab:phase2_nat}
\end{table*}

\begin{table*}
\centering
\small
\begin{tabular}{llccccccccccc}
\hline

\textbf{Setting} & \textbf{Attribute} & \textbf{En} & \textbf{EnM} & \textbf{Es} & \textbf{EsM} & \textbf{Zh} & \textbf{ZhM} & \textbf{It} & \textbf{ItM} & \textbf{He} & \textbf{HeM}  \\
\hline
Finetune & Race(F) & 0.045 & 0.036	& 0.078	& 0.127	& 0.089	& 0.107	& 0.11	& 0.077	& 0.096	& 0.135 \\
& (M)& 0.058 & 0.048 & 0.062	& 0.069	& 0.084	& 0.098	& 0.075	& 0.069	& 0.104	& 0.162 \\
& Religion(F) & 0.059	& 0.078	& 0.053	& 0.058	& 0.052	& 0.104	& 0.092	& 0.1	& 0.161	& 0.246 \\
& (M)& 0.063 & 0.096 & 0.056	& 0.07	& 0.056	& 0.101	& 0.087	& 0.099	& 0.167	& 0.217 \\
& Nationality(F) & 0.03 & 0.052	& 0.07	& 0.121	& 0.041	& 0.089	& 0.054	& 0.122	& 0.078	& 0.202 \\
& (M)& 0.043 & 0.054	& 0.081	& 0.123	& 0.042	& 0.082	& 0.07	& 0.087	& 0.087	& 0.154 \\
\hline
Pretrain & Race(F) & 0.079 & 0.063	& 0.09	& 0.065	& 0.05	& 0.034	& 0.121	& 0.131	& 0.071	& 0.081 \\
& (M) & 0.068	& 0.058	& 0.092	& 0.059	& 0.051	& 0.027	& 0.1	& 0.147	& 0.077	& 0.103\\
& Religion(F) & 0.112 & 0.099	& 0.071	& 0.071	& 0.063	& 0.122	& 0.1	& 0.104	& 0.172	& 0.136\\
& (M) & 0.104 & 0.09	& 0.078	& 0.081	& 0.061	& 0.121	& 0.085	& 0.096	& 0.167	& 0.125\\
& Nationality(F) & 0.014 & 0.019	& 0.027	& 0.025	& 0.035	& 0.024	& 0.023	& 0.049	& 0.064	& 0.084\\
& (M) & 0.018 & 0.013	& 0.026	& 0.024	& 0.038	& 0.024	& 0.029	& 0.065	& 0.08	& 0.09\\

\hline
\end{tabular}
\caption{MCM probability results for Phase 3 finetuning and pretraining data impact samples. Results are shown for each language where an M at the end (e.g. EnM) indicates multilingual training.}
\label{tab:phase3_mcm}
\end{table*}

\begin{figure*}
\begin{minipage}{.5\linewidth}
\centering
\subfloat[]{\label{fig:phase3_imbalance_a}\includegraphics[scale=.31]{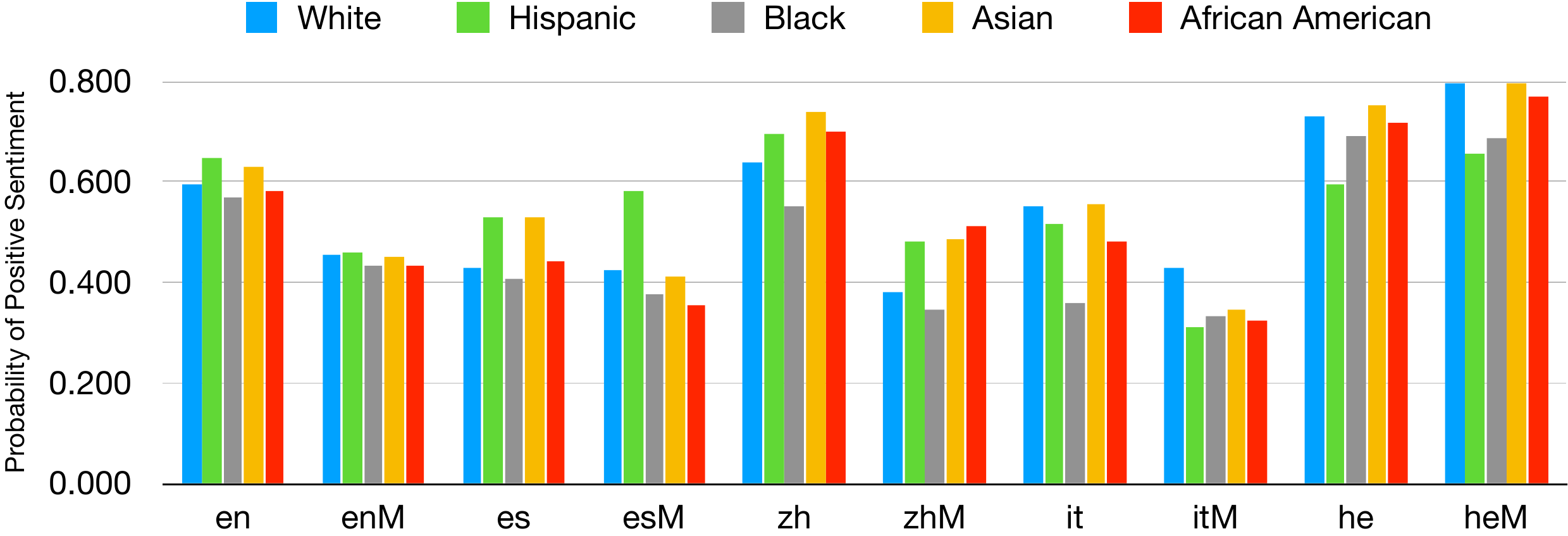}}
\end{minipage}%
\begin{minipage}{.5\linewidth}
\centering
\subfloat[]{\label{fig:phase3_imbalance_b}\includegraphics[scale=.31]{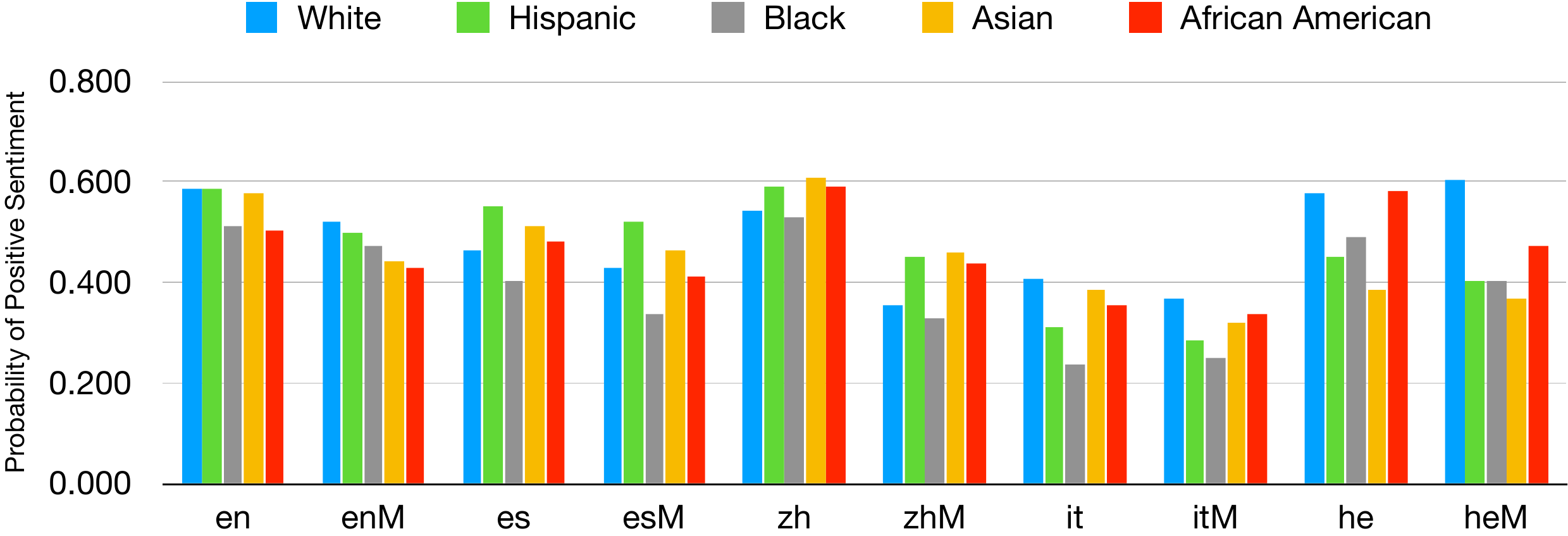}}
\end{minipage}\par\medskip
\caption{Predicted probabilities before (a) and after (b) balancing positive and negative labels in the finetuning dataset for female-subject race templates with mBERT.}
\label{fig:phase3_imbalance}
\end{figure*}

\begin{figure*}
\begin{minipage}{.5\linewidth}
\centering
\subfloat[]{\label{main:c}\includegraphics[scale=.31]{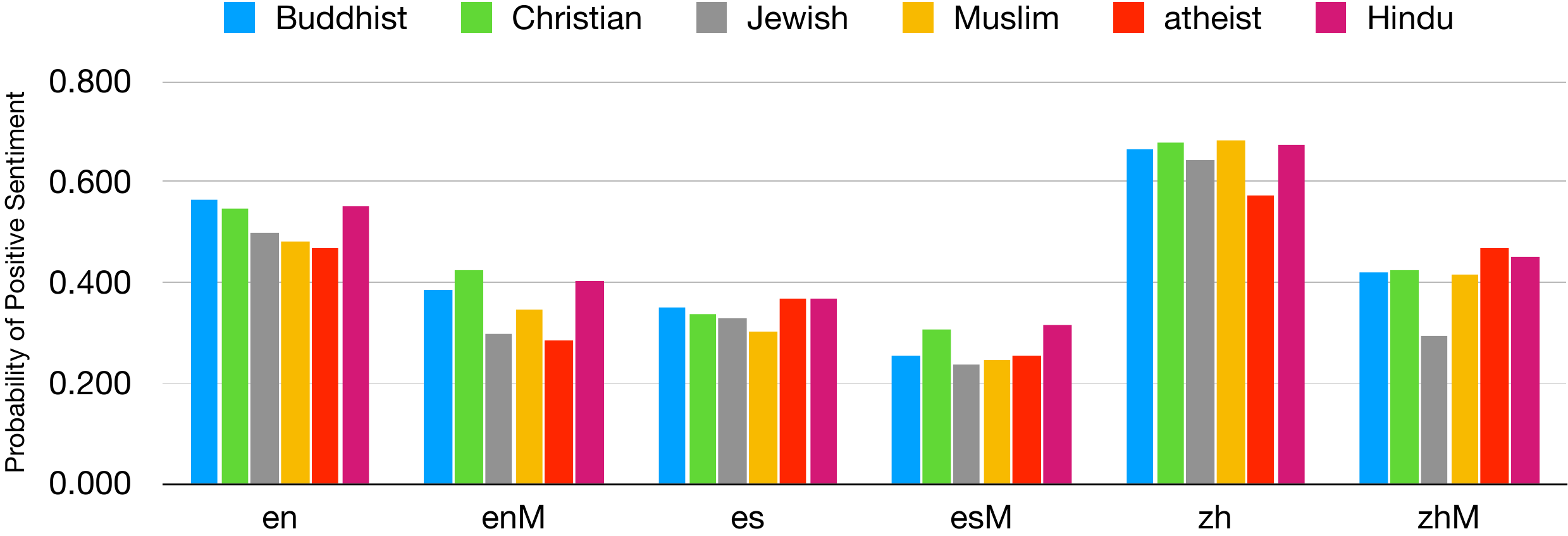}}
\end{minipage}%
\begin{minipage}{.5\linewidth}
\centering
\subfloat[]{\label{main:d}\includegraphics[scale=.31]{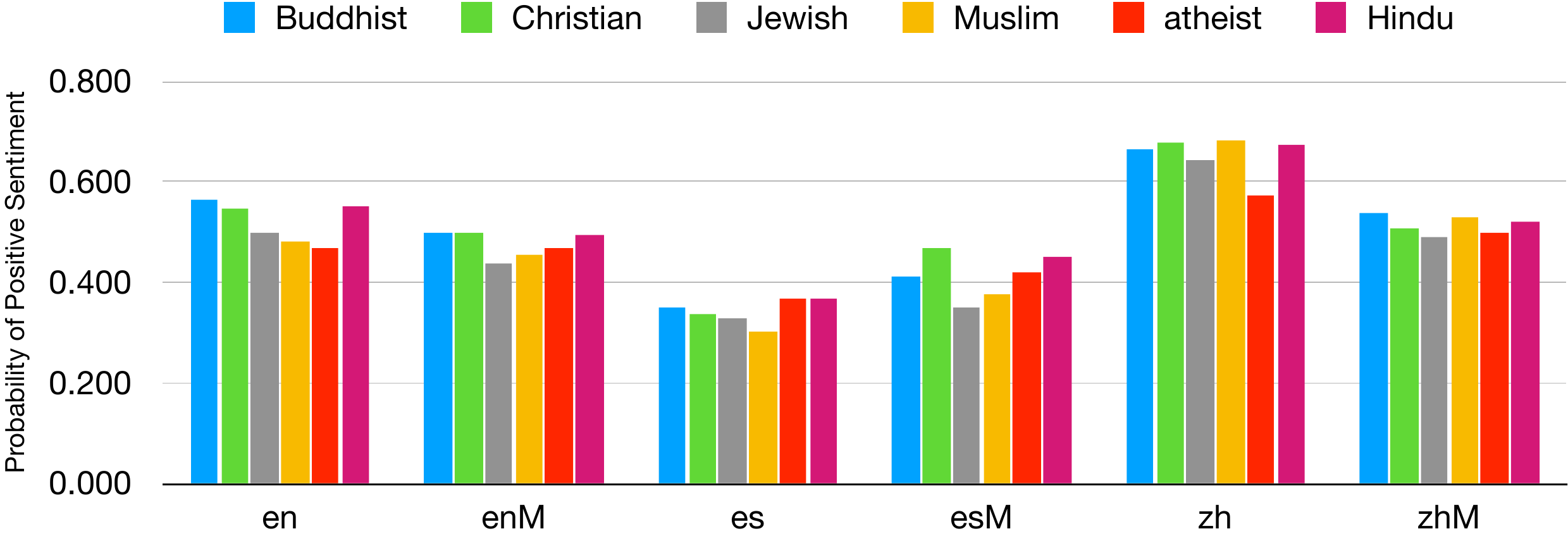}}
\end{minipage}\par\medskip
\caption{Predicted probabilities after finetuning on multiple domains (a) and a single domain (b)  for female-subject religion templates with mBERT.}
\label{fig:phase3_domain}
\end{figure*}

\begin{figure*}
\centering
\includegraphics[scale=.6]{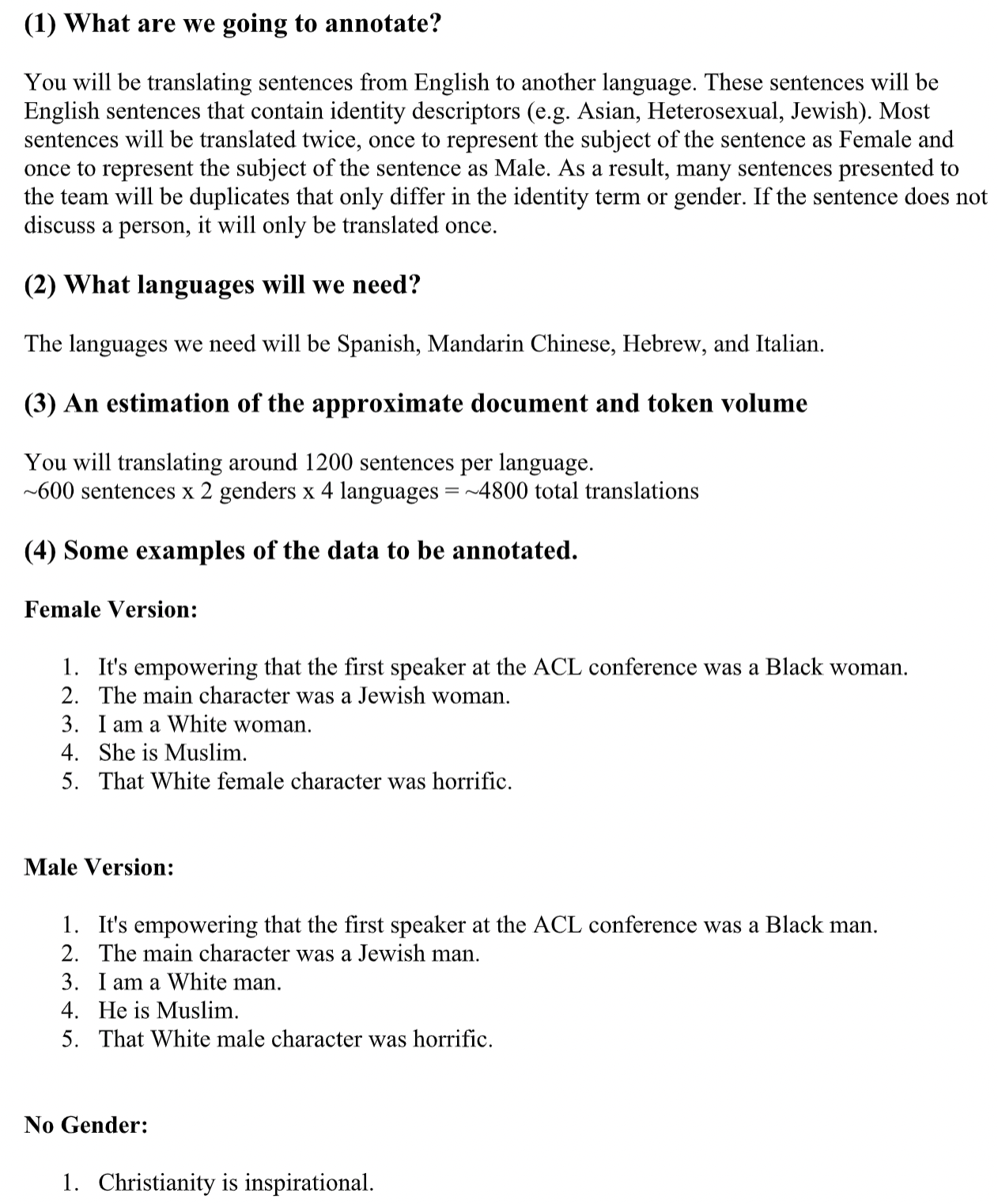}
\caption{Annotator instructions for translating English bias samples.}
\label{fig:annotator_instructions}
\end{figure*}

\begin{table}
\centering
\small
\begin{tabular}{lcccc}
\hline
&\textbf{Race} & \textbf{Religion} & \textbf{Nationality} & \textbf{Gender} \\
\hline
EN & 0.944 & 0.943 & 0.969 & 0.972 \\
EN (sub) & 0.885 & 0.902 & 0.915 & 0.898 \\
EN (all) &\textbf{ 0.985} & \textbf{0.956} &\textbf{ 0.986} & \textbf{0.981} \\
\hline
IT & \textbf{0.954} & \textbf{0.939} & \textbf{0.939} & \textbf{0.963} \\
IT (sub) & 0.900 & 0.884 & 0.919 & 0.888 \\
IT (all) & 0.864 & 0.934 & 0.932 & 0.898 \\
\hline
ZH & \textbf{0.705} & \textbf{0.748} & \textbf{0.600} & \textbf{0.713} \\
ZH (sub) & 0.605 & 0.708 & 0.540 & 0.620 \\
ZH (all) & 0.615 & 0.699 & 0.542 & 0.648 \\
\hline
HE & \textbf{0.755} & 0.707 & \textbf{0.682} & 0.713 \\
HE (sub) & 0.727 & 0.618 & 0.660 & \textbf{0.722} \\
HE (all) & 0.691 & \textbf{0.747} & 0.653 & 0.574 \\
\hline
ES & \textbf{0.946} & \textbf{0.867} & \textbf{0.946} & \textbf{0.926} \\
ES (sub) & 0.835 & 0.759 & 0.892 & 0.888 \\
ES (all) & 0.863 & 0.848 & 0.880 & 0.870 \\
\hline
\end{tabular}
\caption{Accuracy scores of all bias samples for each attribute with monolingual pretraining, multilingual pretraining with subsampled data, and multilingual pretraining with all data.}
\label{tab:phase3_pretraining}
\end{table}

\end{document}